\documentclass{llncs}
\usepackage{llncsdoc}

\usepackage{microtype}
\usepackage{tipa}
\usepackage{mathtools}
\usepackage{fancyref}
\usepackage{float}
\restylefloat{figure}
\usepackage{caption}
\usepackage{amsmath}
\usepackage{amssymb}  
\usepackage{mathtools}
\usepackage{fancyhdr}
\usepackage{kantlipsum}
\usepackage{eso-pic}
\usepackage{nomencl}
\usepackage{etoolbox}
\usepackage{balance}
\usepackage{dsfont}
\usepackage{tikz}
\usepackage{cite}
\usepackage{eucal}
\usepackage{nccmath}
\tikzset{>=latex}
\usetikzlibrary{automata,positioning}
\usepackage[algo2e,linesnumbered,ruled,lined]{algorithm2e}
\usepackage[toc,page]{appendix}
\usepackage{hyperref}
\hypersetup{
    colorlinks=true,
    linkcolor=black,
    filecolor=black,      
    urlcolor=black,
    citecolor=black,
}

\newtheorem{problem_1}{Problem}

\title{Logically-Constrained Reinforcement Learning
}
\author{Mohammadhosein Hasanbeig, Alessandro Abate, \and Daniel Kroening}

\institute{University of Oxford}

\begin{document}

\maketitle

\begin{abstract}
We present the first model-free Reinforcement Learning (RL) algorithm to synthesise policies for an unknown Markov Decision Process (MDP), such that a linear time property is satisfied. The given temporal property is converted into a Limit Deterministic B\"uchi Automaton (LDBA) and a robust reward function is defined over the state-action pairs of the MDP according to the resulting LDBA. With this reward function, the policy synthesis procedure is ``constrained'' by the given specification. These constraints guide the MDP exploration so as to minimize the solution time by only considering the portion of the MDP that is relevant to satisfaction of the LTL property. This improves performance and scalability of the proposed method by avoiding an exhaustive update over the whole state space while the efficiency of standard methods such as dynamic programming is hindered by excessive memory requirements, caused by the need to store a full-model in memory. Additionally, we show that the RL procedure sets up a local value iteration method to efficiently calculate the maximum probability of satisfying the given property, at any given state of the MDP. We prove that our algorithm is guaranteed to find a policy whose traces probabilistically satisfy the LTL property if such a policy exists, and additionally we show that our method produces reasonable control policies even when the LTL property cannot be satisfied. The performance of the algorithm is evaluated via a set of numerical examples. We observe an improvement of one order of magnitude in the number of iterations required for the synthesis compared to existing approaches.
\end{abstract}

\section{Introduction}
\label{sec:intro}

Markov Decision Processes (MDPs), are
discrete-time stochastic control processes that are extensively used for
sequential decision making from automatic control and AI to economics and biology \cite{puterman}. MDPs are suitable for modelling
decision making problems when outcomes of actions are not fully under
the control of a decision maker (or an agent). 
An MDP is said to be solved when the agent is able to select optimal actions, i.e. to come up with an optimal policy at any given state. 

Reinforcement Learning (RL) is a class of machine-learning algorithms that
is widely used to solve MDPs in a variety of applications such as robotics \cite{sutton2,rahili,ifac,thesis}, traffic management \cite{dorsa}, flight control \cite{ng} and human-level game playing \cite{mnih,silver}. Unlike other conventional methods of solving MDPs, such as Dynamic Programming (DP), in an RL setup the graph structure and the stochastic behaviour of the MDP is not necessarily known: the agent \emph{learns} an optimal policy just by interacting with the MDP \cite{otterlo}. Learning via interaction in RL is accomplished in two different ways: model-based learning and model-free learning. Model-based RL attempts to model the MDP, and then based on that model, finds the most appropriate policy. This method is particularly effective if we have an appropriate model that can fit the problem and hence, can be iteratively refined by interacting with the MDP. The second method, model-free RL, learns an optimal policy directly by mapping the state-action pairs to their expected values, relaxing the need for a fitting model. In this sense, model-free RL is more applicable than model-based RL. 

\textbf{Contributions:} In this paper we propose the first model-free RL algorithm to synthesise a control policy for an MDP such that the generated traces satisfy a Linear Temporal Logic (LTL) property. Additionally, we show that maximizing the expected reward in RL is equivalent to maximizing the probability of satisfying the assigned LTL property and we quantify this probability via a value-iteration-based method. Finally, through theoretical results, we show that when the probability of satisfying the LTL property is zero, our algorithm can still produce reasonable policies.

LTL, a modal logic, can express time-dependent logical properties such as safety and liveness \cite{pnueli}. LTL allows one to specify complex (e.g., temporal, sequential, conditional) mission tasks that are hard (if at all possible) to express and to achieve by conventional methods in classical RL, e.g. \cite{sutton,smith}. An LTL property can be expressed via an automaton, namely a finite-state machine \cite{bible}. In general however, LTL-to-automaton translation may result in a non-deterministic automaton, with which probabilistic model checking and policy synthesis for MDPs are in general not semantically meaningful. A standard solution to this issue is to use Safra construction to determinise the automaton, which as expected can increase the size of the automaton dramatically \cite{safra,nba2d}. An alternative solution is to directly convert a given LTL formula into a Deterministic Rabin Automaton (DRA), which by definition is not non-deterministic. Nevertheless, it is known that such conversion results, in the worst case, in automata that are doubly exponential in size of the original LTL formula \cite{dra4}. Although a fully non-deterministic automaton cannot be used for probabilistic model checking, it is known that restricted forms of non-determinism might be allowed: in this work we propose to convert the given LTL property into a Limit Deterministic B\"uchi Automaton (LDBA) \cite{sickert}. 
It is shown that this construction results in an exponential-sized automaton for LTL$\setminus$GU (a fragment of LTL), and results in nearly the same size as a DRA for the remaining class of LTL formulae. Furthermore, the resulting B\"uchi automaton is more succinct than a Rabin automaton in terms of its acceptance conditions, which makes its later use for synthesis much simpler \cite{tkachev}. 

Once the LDBA is generated from the LTL property, we construct an on-the-fly\footnote{On-the-fly means that the algorithm tracks (or executes) the state of a structure (or a function) without explicitly build that structure (or interrupting the run).} and synchronous
product between the MDP and the resulting LDBA and then assign
a reward function to the state-action pairs of the MDP such that it is concurrent with the accepting condition of the LDBA. Using this reward shaping, RL is able to generate a policy
that satisfies the given LTL property with maximum possible probability.

The proposed algorithm is completely ``model-free'', which means that we are able to synthesise policies (1) without knowing the MDP graph and its transition probabilities (as opposed to DP), and (2) without preprocessing or approximating the MDP (unlike model-based RL). The second feature of the proposed algorithm is quite important as it allows the algorithm to synthesise policies directly from the state space of the MDP, which subsequently speeds up the synthesis procedure. 

In parallel to strategy synthesis, we employ an on-the-fly value iteration method to calculate the probability of satisfaction of the LTL property. The use of RL for policy generation allows the value iteration algorithm to focus on parts of the state space that are relevant to the property of interest. This results in a faster calculation of satisfaction probability values when compared to DP, where in contrast these probabilities are computed globally, that is over the whole state space.

\textbf{Related Work:}
The problem of control synthesis for temporal logic has been considered in various works. In \cite{wolf}, the property of interest is LTL, which is converted to a DRA using standard methods. A product MDP is then constructed with the resulting DRA and a modified DP is applied over the product MDP, maximising the worst-case probability of satisfying the specification over all transition probabilities. However, in this work the MDP must be known a priori. Classical DP based algorithms are of limited utility in this context, both because of assumption of a perfect model and because of its great computational costs \cite{sutton}. \cite{topku} assumes that the given MDP model has unknown transition probabilities and builds a Probably Approximately Correct MDP (PAC MDP), which is multiplied by the logical property after conversion to DRA. The goal is to calculate the finite-horizon $T$-step value function for each state such that the value is within an error bound of the actual state value where the value is the probability of satisfying the given LTL property. The PAC MDP is generated via an RL-like algorithm and then value iteration is applied to calculate the values of states. 

The problem of policy generation by maximising the probability of satisfying a given specification for unbounded reachability properties is investigated in \cite{brazdil}. The policy generation in this work relies on approximate DP even when MDP transition probabilities are unknown. This requires a mechanism to approximate these probabilities (much like PAC MDP), and the optimality of the generated policy highly depends on the accuracy of this approximation. Therefore, as stated in \cite{brazdil}, a sufficiently large number of simulations has to be executed to make sure that the probability approximations are accurate enough. Furthermore, the algorithm in \cite{brazdil} assumes prior knowledge about the minimum transition probability. Via LTL-to-DRA conversion, \cite{brazdil} algorithm can be extended to the problem of control synthesis for LTL specifications with the expense of automaton double exponential blow-up. In the same direction, \cite{dorsa} employed a learning based approach to generate a policy that is able to satisfy a given LTL property. However, LTL-to-DRA conversion is in general known to result in large product-MDPs, and also complicated reward shaping due to Rabin accepting condition. Same as \cite{brazdil}, the algorithm in \cite{dorsa} hinges on approximating the transition probabilities which hinders the policy generation process. Additionally, the proposed method in \cite{dorsa} is proved to find only policies that satisfy the given LTL property with probability one, i.e. ``proper policies''. 

Comparing to the above mentioned methods, the proposed algorithm in this work learns the dynamics of the MDP and the optimal policy at the same time, without explicitly approximating the transition probabilities. Moreover, it is able to find the optimal policy even if the probability of satisfaction is not one. In this sense this is the first work on model-free constrained RL and thus we use classical RL as the core of our algorithm. Although the proposed framework is extendible to more recent developments in RL community, e.g. \cite{deepql,asynchronous}, we believe that this is out of the scope of this paper. We provide empirical comparison with these related works later in the paper.

Moving away from RL and full LTL, the problem of synthesising a policy that satisfies a temporal logic specification and that at the same time optimises a performance criterion is considered in \cite{belta2,game}. In \cite{game}, the authors separate the problem into two sub-problems: extracting a (maximally) permissive strategy for the agent and then quantifying the performance criterion as a reward function and computing an optimal strategy for the agent within the operating envelope allowed by the permissive strategy. Similarly, \cite{nils} first computes safe, permissive strategies with respect to a reachability property. Then, under these constrained strategies, RL is applied to synthesise a policy that satisfies an expected cost criterion.  

Truncated LTL is proposed in \cite{belta} as the specification language, and a policy search method is used for synthesis. In \cite{scltl}, scLTL is proposed for mission specification, which results in deterministic finite automata. A product MDP is then constructed and a linear programming solver is used to find optimal policies. PCTL specifications are also investigated in \cite{morteza}, where a linear optimisation solution is used to synthesise a control policy. In \cite{pmc}, an automated method is proposed to verify and repair the policies that are generated by RL with respect to a PCTL formula - the key engine runs by feeding the Markov chain induced by the policy to a probabilistic model checker. In \cite{andersson}, some practical challenges of RL are addressed by letting the agent plan ahead in real time using constrained optimisation. 

The concept of shielding is employed in \cite{shield} to synthesise a reactive system that ensures that the agent stays safe during and after learning. This approach is closely related to teacher-guided RL \cite{teacher}, since a shield can be considered as a teacher, which provides safe actions only if absolutely necessary. However, unlike our focus on full LTL expressivity, \cite{shield} adopted the safety fragment of LTL as the specification language. To express the specification, \cite{shield} uses DFAs and then translates the problem into a safety game. The game is played by the environment and the agent. In every state of the game, the environment chooses an input, and then the agent chooses some output. The game is won by the agent if only safe states are visited during the play. However, the generated policy always needs the shield to be online, as the shield maps every unsafe action to a safe action. 

\textbf{Organization:}
The organisation of the material in this paper is as follows: Section~\ref{background}
reviews basic concepts and definitions. In Section~\ref{cps}, we
discuss the policy synthesis problem and we propose a method to constrain
it. In Section~\ref{psp}, we discuss an online value iteration method to
calculate the maximum probability of satisfying the LTL property at any given
state. Case studies are provided in Section~\ref{case study} to
quantify the performance of the proposed algorithms. 
The Appendix includes proofs of theorems and propositions. Extra material on this paper, including videos of the experiments, can be found at
\begin{center} { \href{https://www.cs.ox.ac.uk/conferences/lcrl/}{cs.ox.ac.uk/conferences/lcrl}} \end{center} 
\section{Background}
\label{background}
%
%
\begin{definition} [Markov Decision Process (MDP)] 
	An MDP $\textbf{M}=(\allowbreak
	\mathcal{S},\allowbreak\mathcal{A},\allowbreak s_0,\allowbreak
	P,\allowbreak\mathcal{AP},\allowbreak L)$ is a tuple over a finite set of
	states $\mathcal{S}$ where $\mathcal{A}$ is a finite set of actions, $s_0$
	is the initial state and $P:\mathcal{S} \times \mathcal{A} \times
	\mathcal{S} \rightarrow [0,1]$ is the transition probability function which determines probability of moving from a given state to another by taking an action. 
	$\mathcal{AP}$ is a finite set of atomic propositions and a labelling
	function $L: \mathcal{S} \rightarrow 2^{\mathcal{AP}}$ assigns to each state
	$s \in \mathcal{S}$ a set of atomic propositions $L(s) \subseteq
	2^\mathcal{AP}$. We use $s   \xrightarrow{a}   s'$ to
	denote a transition from state $s \in \mathcal{S}$ to state $s' \in
	\mathcal{S}$ by action $a \in \mathcal{A}$.$\hfill \lrcorner$
\end{definition}
\begin{definition}
	[Stationary Policy] A stationary randomized policy $\mathit{Pol}: \mathcal{S} \times \mathcal{A} \rightarrow [0,1]$ is a mapping from each state $s \in \mathcal{S}$, and action $a \in \mathcal{A}$ to the probability of taking action $a$ in state $s$. A deterministic policy is a degenerate case of a randomized policy which outputs a single action at given state, that is $\exists a\in\mathcal{A}$ such that $\mathit{Pol}(s,a)=1$.$\hfill \lrcorner$
\end{definition}

Let the MDP $\textbf{M}$ be a model that formulates the interaction between
the agent and its environment. We define a function $R:\mathcal{S}\times\mathcal{A}\rightarrow \mathds{R}_0^+$ that returns immediate reward received by the agent after performing action $a \in \mathcal{A}$ in state $s
\in \mathcal{S}$. In an RL setup, the reward function specifies what agent needs to achieve, not how to achieve it. We would like the agent itself to produce an optimal solution.

Q-learning (QL), is the most widely-used RL algorithm for solving MDPs from simple control tasks \cite{sutton} to more recent developments that go beyond human-level control skills \cite{deepql}. 

For each state $s \in \mathcal{S}$ and for any available
action $a \in \mathcal{A}_s$, QL assigns a quantitative value
$Q:\mathcal{S}\times\mathcal{A}\rightarrow \mathds{R}$, which is initialized
with an arbitrary fixed value for all state-action pairs. As the agent
starts learning and receiving rewards, the $Q$ function is updated by the
following rule:
\[\left\{
\begin{array}{lr}
Q(s,a) \leftarrow Q(s,a)+\mu
[R(s,a)+\gamma \max\limits_{a' \in \mathcal{A}_s}(Q(s',a'))-Q(s,a)],\\
Q(s'',a'') \leftarrow Q(s'',a''), 
\end{array}
\right.
\]
where $ Q(s,a) $ is the $Q$-value corresponding to state-action $ (s,a) $, $
\mu\in (0,1] $ is the step size or learning rate, $ R(s,a) $ is the agent realized reward for
performing action $a$ in state $s$, $\gamma \in [0,1)$ is a coefficient called
discount factor, $s'$ is the state after performing action $a$, and $(s'',a'')$ refers to any state-action pair other than $ (s,a) $. Note that calculation of $Q$-values does not require any knowledge or approximation of the MDP transition probabilities.$\hfill \lrcorner$

\begin{definition}
	[Expected Discounted Utility] \label{expectedut} For a policy $\mathit{Pol}$ on an MDP $\textbf{M}$, the expected discounted utility is defined as \cite{sutton}:
	$$
	{U}^{\mathit{Pol}}(s)=\mathds{E}^{\mathit{Pol}} [\sum\limits_{n=0}^{\infty} \gamma^n~ R(s_n,Pol(s_n))|s_0=s],
	$$
	where $\mathds{E}^{\mathit{Pol}} [\cdot]$ denotes the expected value given that the agent follows policy $\mathit{Pol}$, $\gamma$ is the discount factor, and $s_0,...,s_n$ is the sequence of states generated by policy $\mathit{Pol}$ up to time step $n$. $\hfill \lrcorner$
\end{definition}
\begin{definition}[Optimal Policy]\label{optimal_pol}
	Optimal policy $\mathit{Pol}^*$ is defined as follows:
	$$
	\mathit{Pol}^*(s)=\arg\max\limits_{Pol \in \mathcal{D}}~ {U}^{\mathit{Pol}}(s),
	$$
	where $\mathcal{D}$ is the set of all stationary deterministic policies over $\mathcal{S}$.$\hfill\lrcorner$
\end{definition}

\begin{theorem}
	In any finite-state MDP $\textbf{M}$, if there exists an optimal policy, then that policy is stationary and deterministic \cite{puterman}.$\hfill\lrcorner$
\end{theorem}

It has been proven \cite{watkins} that $Q$ values in QL algorithm converge to $Q^*$ such that ${U}^{\mathit{Pol}^*}(s)=\max\limits_{a \in \mathcal{A}_s} Q^*(s,a)$. From Definition \ref{expectedut} it is easy to see:

\begin{equation}
\label{qnv}
Q^*(s,a)=R(s,a)+\gamma\sum\limits_{s'\in\mathcal{S}} P(s,a,s') U^{\mathit{Pol}^*}(s').
\end{equation}

Now suppose that the agent is in state $s$. The simplest method to select an optimal action at each state $s$ (i.e. to
synthesize an optimal policy) is to choose an action that yields the highest
$Q$-value at each state (i.e. a greedy action selection). Once QL converges to $ Q^* $ then the optimal policy
$\mathit{Pol}^*: \mathcal{S} \rightarrow \mathcal{A}$ can be generated as:
$$
\mathit{Pol}^*(s)=\arg\max\limits_{a \in \mathcal{A}}~Q^*(s,a).
$$
%

\section{Constrained Policy Synthesis}
\label{cps}

So far, we discussed how the agent can achieve the desired objective by
following a reward function over the states of the MDP. The reward function
can guide the agent to behave in a certain way during learning. In order
to specify a set of desirable constraints (i.e. properties) over the
agent evolution we employ Linear Temporal Logic (LTL). The
set of LTL formulae over the set of atomic proposition $\mathcal{AP}$ is
defined as
\begin{equation}
\label{ltlsyntax}
\varphi::= true ~|~ \alpha \in \mathcal{AP} ~|~ \varphi \land \varphi ~|~ \neg \varphi ~|~ \bigcirc \varphi ~|~ \varphi \cup \varphi.
\end{equation}
In the following we define LTL formulae interpreted over MDPs.
\begin{definition}
[Path] \label{run}
In an MDP $\textbf{M}$, an infinite path $\rho$ starting at $s_0$ is a
sequence of states $\rho= s_0 \xrightarrow{a_0} s_1 \xrightarrow{a_1} ... ~$ such that every transition $s_i \xrightarrow{a_i} s_{i+1}$ is possible in $\textbf{M}$,
i.e. $P(s_i,a_i,s_{i+1})>0$. We might also denote $\rho$ as $s_0..$ to emphasize that $\rho$ starts from $s_0$. A finite path is a
finite prefix of an infinite path.~\hfill~$\lrcorner$
\end{definition}
\vspace{1mm}
Given a path $\rho$, the $i$-th state of $\rho$ is denoted by $\rho[i]$
where $\rho[i]=s_{i}$. Furthermore, the $i$-th suffix of $\rho$ is
$\rho[i..]$ where $\rho[i..]=s_i \rightarrow s_{i+1} \rightarrow s_{i+2}
\rightarrow s_{i+3} \rightarrow ...~$.

\begin{definition}
[LTL Semantics] \label{syntax} 
For an LTL formula $\varphi$ and for a path $\rho$, the satisfaction relation $\rho\models\varphi$ is defined as
\begin{equation*}
\begin{aligned}
& \rho \models \alpha \in \mathcal{AP} \iff \alpha \in L(\rho[0]), \\
& \rho \models \varphi_1\wedge \varphi_2 \iff \rho \models \varphi_1\wedge \rho \models \varphi_2,\\
& \rho \models \neg \varphi \iff \rho \not \models \varphi, \\
& \rho \models \bigcirc \varphi \iff \rho[1..] \models \varphi, \\
& \rho \models \varphi_1\cup \varphi_2 \iff \exists j \in \mathds{N}\cup\{0\} ~\mbox{s.t.}~ \rho[j..] \models \varphi_2 ~and \\ 
& \hspace{3cm}\forall i,~0 \leq i < j,~ \rho[i..] \models \varphi_1.
\end{aligned}
\end{equation*} 
\hfill $\lrcorner$
\end{definition}

Using the until operator we are able to define two temporal modalities: (1)
eventually, $\lozenge \varphi = true \cup \varphi$; and (2) always, $\square
\varphi = \neg \lozenge \neg \varphi$. An LTL formula $\varphi$ over
$\mathcal{AP}$ specifies the following set of words:
$$
\mathit{Words}(\varphi)=\{\sigma \in (2^{\mathcal{AP}})^\omega ~\mbox{s.t.}~ \sigma \models \varphi\}.
$$

\begin{definition}
[Probability of Satisfying an LTL Formula] \label{ltlprobab}
Starting from $s_0$, we define the probability of satisfying formula $\varphi$ as
$$
\mathit{Pr}(s_0..^{\mathit{Pol}} \models \varphi),
$$
where $s_0..^{\mathit{Pol}}$ is the collection of all paths starting from $ s_0 $, generated by policy $\mathit{Pol}: \mathcal{S} \rightarrow \mathcal{A}$.~\hfill$\lrcorner$
\end{definition}
\begin{definition}[Policy Satisfaction]
	We say that a stationary deterministic policy $ \mathit{Pol} $ satisfies an LTL formula $ \varphi $ if: 
	$$
	\mathit{Pr}(s_0..^{\mathit{Pol}} \models \varphi)>0,
	$$
	where $ s_0 $ is the initial state of the MDP.
	\hfill $\lrcorner$ 
\end{definition}

Using an LTL formula we can specify a set of constrains (i.e. properties) over the
sequence of states that are generated by the policy in the MDP. Once a policy is selected (e.g. $Pol$) then at each state of an MDP
$\textbf{M}$, it is clear which action has to be taken. Hence, the MDP
$\textbf{M}$ is reduced to a Markov chain, which we denote
by~$\textbf{M}^{\mathit{Pol}}$. 

For an LTL
formula $\varphi$, an alternative method to express the set $Words(\varphi)$ is to employ an LDBA \cite{sickert}. We need to first define a Generalized B\"uchi Automaton (GBA) and then we formally introduce an LDBA \cite{sickert}.

\begin{definition}
[Generalized B\"uchi Automaton] 
A GBA $\textbf{N}=(\allowbreak\mathcal{Q},\allowbreak q_0,\allowbreak\Sigma, \allowbreak\mathcal{F}, \allowbreak\Delta)$ is a structure where $\mathcal{Q}$ is a finite set of states, $q_0 \in \mathcal{Q}$ is the initial state, $\Sigma=2^{\mathcal{AP}}$ is a finite alphabet, $\mathcal{F}=\{F_1,...,F_f\}$ is the set of accepting conditions where $F_j \subset \mathcal{Q}, 1\leq j\leq f$, and $\Delta: \mathcal{Q} \times \Sigma \rightarrow 2^\mathcal{Q}$ is a transition relation. \hfill $\lrcorner$ 
\end{definition}
Let $\Sigma^\omega$ be the set of all
infinite words over $\Sigma$. An infinite word $w \in \Sigma^\omega$ is
accepted by a GBA $\textbf{N}$ if there exists an infinite run $\theta \in
\mathcal{Q}^\omega$ starting from $q_0$ where $\theta[i+1] \in
\Delta(\theta[i],\omega[i]),~i \geq 0$ and for each $F_j \in \mathcal{F}$
\begin{equation} \label{acc}
\mathit{inf}(\theta) \cap F_j \neq \emptyset,
\end{equation}
where $\mathit{inf}(\theta)$ is the set of states that are visited
infinitely often in the sequence $\theta$. The accepted language of the
GBA~$\textbf{N}$ is the set of all infinite words accepted by the
GBA~$\textbf{N}$ and it is denoted by $\mathcal{L}_\omega(\textbf{N})$.

\begin{definition}
[LDBA] \label{ldbadef}
A GBA $\textbf{N}=(\mathcal{Q},q_0,\Sigma, \mathcal{F}, \Delta)$ is
limit deterministic if $\mathcal{Q}$ can be partitioned into two disjoint sets $\mathcal{Q}=\mathcal{Q}_N \cup \mathcal{Q}_D$, such that \cite{sickert}:
\begin{itemize}
\item $\Delta(q,\alpha) \subset \mathcal{Q}_D$ and $|\Delta(q,\alpha)|=1$ for every state $q\in\mathcal{Q}_D$ and for every $\alpha \in \Sigma$,
\item for every $F_j \in \mathcal{F}$, $F_j \subset \mathcal{Q}_D$. \hfill$\lrcorner$
\end{itemize}
\end{definition}

\begin{remark}
The LTL-to-LDBA algorithm proposed in \cite{sickert} (and used in this paper) results in an automaton with two parts: initial ($\mathcal{Q}_N$) and accepting ($\mathcal{Q}_D$). The accepting set is invariant which means that when $\mathcal{Q}_D$ is reached the automaton cannot escape from $\mathcal{Q}_D$. Both the initial part and the accepting part are deterministic and there are non-deterministic $\varepsilon$-transitions between the initial part and the accepting part. An $\varepsilon$-transition allows an automaton to change its state spontaneously and without reading an input symbol. The $\varepsilon$-transitions between $ \mathcal{Q}_N $ and $ \mathcal{Q}_D $ reflect the automaton guesses on reaching $ \mathcal{Q}_D $. Therefore, if after an $\varepsilon$-transition the associated labels cannot be read or the accepting states cannot be visited, then the guess was wrong and the trace is disregarded. Note that according to Definition \ref{ldbadef} the discussed structure is still a limit-deterministic, the determinism in the initial part is stronger than that required in the definition.$\hfill \lrcorner$
\end{remark}

\begin{problem_1}\label{prob1}\hspace{-2mm}\textbf{.}
We are interested in synthesizing an optimal policy for an unknown MDP such that the resulting policy satisfies a given LTL property. 
\end{problem_1} 

To explain core ideas of the algorithm and for ease of exposition only, for now, we assume that the MDP graph and the associated transition probabilities are known. Later, when we introduce our algorithm these assumptions are entirely removed. 

We first propose to generate an LDBA from the desired LTL property and then construct a synchronised structure between the (yet assumed to be known) MDP and the B\"uchi automaton, defined as follows. 

\begin{definition}
[Product MDP] \label{product_mdp_def}
Given an MDP $\textbf{M}=(\mathcal{S},\mathcal{A},s_0,P,\mathcal{AP},L)$ and
an LDBA $\textbf{N}=(\mathcal{Q},q_0,\Sigma, \mathcal{F}, \Delta)$ with $\Sigma=2^{\mathcal{AP}}$, the product MDP is defined as $\textbf{M}\otimes
\textbf{N} = \textbf{M}_\textbf{N}=(\mathcal{S}^\otimes,\allowbreak
\mathcal{A},\allowbreak s^\otimes_0,P^\otimes,\allowbreak
\mathcal{AP}^\otimes,\allowbreak L^\otimes)$, where $\mathcal{S}^\otimes =
\mathcal{S}\times\mathcal{Q}$, $s^\otimes_0=(s_0,q_0)$,
$\mathcal{AP}^\otimes = \mathcal{Q}$, $L^\otimes =
\mathcal{S}\times\mathcal{Q}\rightarrow 2^\mathcal{Q}$ such that
$L^\otimes(s,q)={q}$, $P^\otimes:\mathcal{S}^\otimes \times \mathcal{A}
\times \mathcal{S}^\otimes \rightarrow [0,1]$ is the transition probability
function such that $(s_i \!  \xrightarrow{a} \!  s_j) \wedge (q_i \! 
\xrightarrow{L(s_j)} \!  q_j) \Rightarrow
P^\otimes((s_i,q_i),a,(s_j,q_j))=P(s_i,a,s_j)$. Over the states of the product MDP we also define accepting condition $\mathcal{F}^\otimes=\{F^\otimes_1,...,F^\otimes_f\}$ where ${F}^\otimes_j=\mathcal{S}\times F_j$. 
\hfill $\lrcorner$
\end{definition}

\begin{remark}
In order to handle $\varepsilon$ transitions in the constructed LDBA we have to add the following modifications to the standard definition of the product MDP:
\begin{itemize}
\item for every potential $\varepsilon$-transition to some state $q \in \mathcal{Q}$ we add a corresponding
action $\varepsilon_q$ in the product:
$$
\mathcal{A}^\otimes=\mathcal{A}\cup \{\varepsilon_q, q \in \mathcal{Q}\}.
$$
\item The transition probabilities of $\varepsilon$-transitions are given by 
\[P^\otimes((s_i,q_i),\varepsilon_q,(s_j,q_j)) = \left\{
  \begin{array}{lr}
    1 & $ if $  s_i=s_j,~q_i\xrightarrow{\varepsilon_q} q_j=q,\\
    0 & $ otherwise. $
\end{array}
\right.
\]
\end{itemize}
$\hfill \lrcorner$
\end{remark}

The resulting product is an MDP over which we can define a reward
function and run the QL algorithm. Intuitively, the product MDP is a synchronous structure encompassing
transition relations of the original MDP and also the structure of the
B\"uchi automaton, and as a result, inherits characteristics of both. Therefore, a
proper reward shaping can lead the RL agent to find a policy that is
optimal and probabilistically satisfies the LTL property~$\varphi$. Before introducing the reward assignment considered in this paper, 
we need to define the ensuing notions. 

%
%

\begin{definition}[Non-accepting Sink Component]
A non-accepting sink component in LDBA $\textbf{N}=(\mathcal{Q},q_0,\Sigma, \mathcal{F}, \Delta)$ is a directed graph induced by a set of states $ Q \subset\mathcal{Q}$ such that (1) is strongly connected, (2) does not include all accepting sets $ F_k,~k=1,...,f $, and (3) there exist no other set $ Q' \subset \mathcal{Q},~Q'\neq Q $ that $ Q \subset Q' $. We denote the union of all non-accepting sink components as \textit{SINKs}. Note that \textit{SINKs} can be empty. \hfill $\lrcorner$
\end{definition}

\begin{definition}
[Accepting Frontier Function] \label{frontier} For an LDBA $\textbf{N}=(\mathcal{Q},q_0,\Sigma,\allowbreak\mathcal{F},\allowbreak\Delta)$, we define the function $ Acc:\mathcal{Q}\times 2^{\mathcal{Q}}\rightarrow2^\mathcal{Q} $ as the accepting frontier function, which executes the following operation over a given set $ \mathds{F}: $ 

\[Acc(q,\mathds{F})=\left\{
\begin{array}{lr}
\mathds{F}\setminus F_j & $ if $  q \in F_j \wedge \mathds{F}\neq F_j,\\
\bigcup\limits_{k=1}^{f} F_k \setminus F_j & $ if $ q \in F_j \wedge \mathds{F}=F_j,\\
\mathds{F} & $otherwise.$
\end{array}
\right.
\] 

Once the state $ q\in F_j $ and the set $ \mathds{F} $ are introduced to the function $ Acc $, it outputs a set containing the elements of $ \mathds{F} $ minus those elements that are common with $ F_j $. However, if by any chance, $ \mathds{F}=F_j $ then the output is union of all accepting sets of the LDBA minus those elements that are common with $ F_j $. If the state $ q $ is not an accepting state then the output of $ Acc $ is $ \mathds{F} $.
\hfill $\lrcorner$
\end{definition}

\begin{remark}
In order to clearly elucidate the role of different components and techniques in our algorithm, we employed notions such as transition probabilities, and product MDP. However, as emphasised before the algorithm can indeed run \textbf{model-free}, and as such does not depend on these model components. As per Definition \ref{ldbadef}, the LDBA is composed of two disjoint sets of states $ \mathcal{Q}_D $ (which is invariant) and $ \mathcal{Q}_N $, where the accepting states belong to the set $ \mathcal{Q}_D $. Since all transitions are deterministic within $ \mathcal{Q}_N $ and $ \mathcal{Q}_D $, the automaton transitions can be executed \textbf{only} by reading the labels, which makes the agent aware of the automaton state without explicitly constructing the product MDP. We will later define a reward function that we will call \textbf{on-the-fly}, to emphasize the fact that the agent does not need to know the structure and transition probabilities of the original MDP (and its product). \hfill $\lrcorner$
\end{remark}

In the following, we propose a reward function that observes the current state $ s^\otimes $, action $ a $, and also the subsequent state $ {s^\otimes}' $. The reward function then gives the agent a scalar reward according to the following rule:
\[R(s^\otimes,a) = \left\{
\begin{array}{lr}
r_p & $ if $  q' \in \mathds{A},~{s^\otimes}'=(s',q'),\\
r_n & $ otherwise.$
\end{array}
\right.
\] 
Here ${s^\otimes}'$ is the state that is probabilistically reached from state ${s^\otimes}$
by taking action $a$, $r_p>0$ is an arbitrary positive reward, and $r_n=0$  is a neutral
reward. The set $ \mathds{A} $ is called the accepting frontier set, initialised as $ \mathds{A}=\bigcup\limits_{k=1}^{f} F_k $ and is updated by the following rule every time after the reward function is evaluated:
$$
\mathds{A}\leftarrow Acc(q',\mathds{A})
$$

Intuitively, $ \mathds{A} $ always contains those accepting states that are needed to be visited at a given time and in this sense the reward function is adaptive to the LDBA accepting condition. Thus, the agent is guided by the above reward assignment to visit these states and once all of the sets $ F_k,~k=1,...,f $ are visited, the accepting frontier $ \mathds{A} $ is reset. In this way the agent is guided to visit the accepting sets infinitely often, and consequently, to satisfy the given LTL property. 

\begin{theorem}
\label{thm}
Let MDP $\textbf{M}_\textbf{N}$ be the product of an MDP~$\textbf{M}$ and an
automaton $\textbf{N}$ where $\textbf{N}$ is the LDBA associated with the
desired LTL property $\varphi$. If a satisfying policy exists then the LCRL optimal policy, which optimizes the expected utility, will find this policy. 
\hfill $\lrcorner$
\end{theorem}

\begin{definition}[Satisfaction Closeness]
Assume that two policies $ \mathit{Pol}_1 $ and $ \mathit{Pol}_2 $ do not satisfy the property $ \varphi $. In other words there are some automaton accepting sets that has no intersection with runs of induced Markov chains $ \mathbf{M}^{\mathit{Pol}_1} $ and $ \mathbf{M}^{\mathit{Pol}_2} $. However, $ \mathit{Pol}_1 $ is closer to property satisfaction if runs of $ \mathbf{M}^{\mathit{Pol}_1} $ has intersection with more automaton accepting sets than runs of $ \mathbf{M}^{\mathit{Pol}_2} $.
\end{definition}

\begin{corollary}
\label{thm2}
If no policy in MDP~$\textbf{M}$ can be generated to satisfy the property $ \varphi $, LCRL is still able to produce the best policy that is closest to satisfying the given LTL formula.
\hfill $\lrcorner$
\end{corollary}

\begin{theorem}
\label{thm1}
If the LTL property is satisfiable by MDP~$\textbf{M}$, then the optimal policy generated by LCRL, maximizes the probability of property satisfaction.~\hfill$\lrcorner$
\end{theorem}

Note that the optimal policy maximizes the expected long-run reward where the only source of reward is the satisfaction of the property. Thus, maximizing the expected reward is equivalent to maximizing the probability of satisfying the LTL formula.

Once the optimal policy $\mathit{Pol}^*:\mathcal{S}^\otimes \rightarrow
\mathcal{A}$ is obtained for the product MDP $\textbf{M}_\textbf{N}$, it
induces a Markov chain, which we denote by
$\textbf{M}_\textbf{N}^\mathit{{Pol}^{*}}$. If we monitor the
traces of states in $\mathcal{S}$, the optimal policy $\mathit{Pol}^*$ also
induces a Markov chain $\mathit{\textbf{M}^{{{Pol}}^{*}}}$
over the original MDP $\textbf{M}$. This induced Markov chain
$\mathit{\textbf{M}^{{{Pol}}^{*}}}$ satisfies the desired LTL
property with probability
\begin{align*}
&\mathit{Pr} (s_0^\otimes..^{\mathit{Pol}^*} \models \varphi)=\mathit{Pr}(\{\rho:~\forall j~ \mathit{inf}(\rho)\cap F^\otimes_j \neq \emptyset \}),
\end{align*}
where $\rho$ is a run of
$\mathit{\textbf{M}_\textbf{N}^{{{Pol}}^{*}}}$ initialized at
$s^\otimes_0$.

So far we discussed a model-free RL algorithm capable of synthesising policies that can respect an LTL formula. In the following we present an optional component of our method that can quantify the quality of the produced policy by calculating the probability of satisfaction associated to the policy. 

\section{Probability of Property Satisfaction}
\label{psp}

The Probability of Satisfaction of a Property (PSP) can be calculated via standard DP, as implemented for instance in PRISM~\cite{prism}. However, as discussed before, DP is quite limited when the state space of the given MDP is large. 

In this section we propose a local value iteration method as part of LCQL that calculates this probability in parallel with the RL scheme.  RL guides the local update of the value iteration, such that it only focuses on parts of the state space that are relevant to the satisfaction of the property. This allows the value iteration to avoid an exhaustive search and thus to converge faster.   

Recall that the transition probability function $ P^\otimes $ is not known. Further, according to Definition \ref{product_mdp_def}, $ P^\otimes((s_i,q_i),a,(s_j,q_j))=P(s_i,a,s_j) $ if $ (s_i \!  \xrightarrow{a} \!  s_j)$ and $(q_i \! \xrightarrow{L(s_j)} \!  q_j) $, showing the intrinsic dependence of $ P^\otimes $ on $ P $. This allows to apply the definition of $ \alpha$-approximation in MDPs \cite{alphaMDP} as follows. 

Let us introduce two functions $ \Psi: \mathcal{S} \times \mathcal{A} \rightarrow \mathds{N}$ and $ \psi: \mathcal{S} \times \mathcal{A} \times \mathcal{S} \rightarrow \mathds{N} $ over the MDP $\textbf{M}$. Function $ \psi(s,a,{s'}) $ represents the number of times the agent executes action $ a $ in state $ s $, thereafter moving to state $ {s}' $, whereas $ \Psi(s,a)=\sum_{{s}' \in \mathcal{S}} \psi(s,a,{s}')$. The maximum likelihood of $ P(s,a,{s}') $ is a Gaussian normal distribution with the mean $ \bar{P}(s,a,{s}')=\psi(s,a,{s}')/\Psi(s,a) $, so that the variance of this distribution asymptotically converges to zero and $ \bar{P}(s,a,{s}')={P}(s,a,{s}') $. Function $ \Psi(s,a) $ is initialised to be equal to one for every state-action pair (reflecting the fact that at any given state it is possible to take any action, and also avoiding division by zero), and function $ \psi(s,a,{s'}) $ is initialised to be equal to zero. 

Consider a function $\mathit{PSP}:\mathcal{S}^\otimes \rightarrow [0,1]$. For a given state $s^\otimes=(s,q)$, the $\mathit{PSP}$ function is initialised as $\mathit{PSP}(s^\otimes)=0$ if $q$ belongs to \textit{SINKs}. Otherwise, it is initialised as $\mathit{PSP}(s^\otimes)=1$. Recall that \textit{SINKs} are those components in the automaton that are surely non-acceptable and impossible to escape from.

\begin{definition}[Optimal PSP]
	The optimal PSP vector is denoted by $\underline{\mathit{PSP}}^*=(\allowbreak\mathit{PSP}^*(s_1),\allowbreak ...,\allowbreak\mathit{PSP}^*(s_{|\mathcal{S}|}))$, where $\mathit{PSP}^*:\mathcal{S}^\otimes \rightarrow [0,1]$ is the optimal PSP function and $\mathit{PSP}^*(s_i)$ is the optimal PSP value starting from state $s_i$ such that
	
	$$
	\mathit{PSP}^*(s_i)=\sup\limits_{\mathit{Pol} \in \mathcal{D}} \mathit{PSP}^{\mathit{Pol}}(s_i),
	$$ 
	where $\mathit{PSP}^{\mathit{Pol}}(s_i)$ is the PSP value of state $s_i$ if we use the policy $\mathit{Pol}$ to determine subsequent states.\hfill$\lrcorner$ 
\end{definition}

In the following we prove that a proposed update rule that makes $\mathit{PSP}$ converge to $\mathit{PSP}^*$.

\begin{definition}
	[Bellman operation \cite{NDP}] For any vector such as $\underline{PSP}=(\allowbreak\mathit{PSP}(s_1),\allowbreak ...,\allowbreak\mathit{PSP}(s_{|\mathcal{S}|}))$ in the MDP $\textbf{M}=(\mathcal{S},\mathcal{A},s_0,P,\mathcal{AP},L)$, the Bellman DP operation $T$ over the elements of $\underline{PSP}$ is defined as:
	
	\begin{equation}
	T~\mathit{PSP}(s_i)=\max\limits_{a \in \mathcal{A}_{s_i}}\sum_{s' \in \mathcal{S}} P(s_i,a,s') \mathit{PSP}(s'). 
	\end{equation} 
	
	If the operation $T$ is applied over all the elements of $\underline{PSP}$, we denote it as $T~\underline{PSP}$.\hfill$\lrcorner$
\end{definition}

\begin{proposition}[From \cite{NDP}]
	The optimal PSP vector $\underline{\mathit{PSP}}^*$ satisfies the following equation:  
	$$
	\underline{\mathit{PSP}}^*=T~\underline{\mathit{PSP}}^*,
	$$	
	and additionally, $\underline{\mathit{PSP}}^*$ is the ``only'' solution of the equation $\underline{PSP}=T~\underline{PSP}
	$.\hfill$\lrcorner$
\end{proposition}

\begin{algorithm2e}[!t]
	\DontPrintSemicolon
	\SetKw{return}{return}
	\SetKwRepeat{Do}{do}{while}
	\SetKwFunction{terminate}{episode$\_$terminate}
	\SetKwFor{terminatedef}{episode$\_$terminate()}{}{}
	\SetKwData{conflict}{conflict}
	\SetKwData{safe}{safe}
	\SetKwData{sat}{sat}
	\SetKwData{unsafe}{unsafe}
	\SetKwData{unknown}{unknown}
	\SetKwData{true}{true}
	\SetKwData{false}{false}
	\SetKwInOut{Input}{input}
	\SetKwInOut{Output}{output}
	\SetKwFor{Loop}{Loop}{}{}
	\SetKw{KwNot}{not}
	\begin{small}
		\Input{LTL specification, $ \textit{it\_threshold} $, $ \gamma $, $ \mu $}
		\Output{$\mathit{Pol}^*$ and $\underline{\mathit{PSP}}^*$}
		initialize $Q: \mathcal{S}^\otimes \times \mathcal{A}^\otimes \rightarrow \mathds{R}^+_0$\;
		initialize $\mathit{PSP}:\mathcal{S}^\otimes\rightarrow [0,1]$\;
		initialize $ \psi: \mathcal{S} \times \mathcal{A} \times \mathcal{S} \rightarrow \mathds{N} $ \;
		initialize $ \Psi: \mathcal{S} \times \mathcal{A} \rightarrow \mathds{N} $ \;
		convert the desired LTL property to an LDBA $\textbf{N}$\;
		initialize $ \mathds{A} $\;
		initialize $episode$-$number:=0$\;
		initialize $iteration$-$number:=0$\;
		\While{$Q$ is not converged}
		{
			$episode$-$number++$\;
			$s^\otimes=(s_0,q_0)$\;
			\While{$ (q \notin {\mathit{SINKs}}:~s^\otimes=(s,q))~ \& ~ (iteration$-$number<\text{it\_threshold})$}
			{
				$iteration$-$number++$\;
				choose $a_*=\mathit{Pol}({s^\otimes})=\arg\max_{a \in \mathcal{A}} Q({s^\otimes},a)$ $ ~\#$ or $ \epsilon-$greedy with diminishing $ \epsilon $\;
				$ \Psi(s^\otimes,a_*)++ $\;
				move to $s^\otimes_*$ by $a_*$\;
				\uIf{$\Psi(s^\otimes,a_*)=2$}
				{
					$ \psi(s^\otimes,a_*,{s^\otimes_*})=2 $
				}
				\Else
				{
					$\psi(s^\otimes,a_*,{s^\otimes_*})++$
				}   
				receive the reward $R({s^\otimes},a_*)$\;
				$\mathds{A}\leftarrow Acc(s_*,\mathds{A})$\;
				$Q({s^\otimes},a_*)\leftarrow Q({s^\otimes},a_*)+\mu[R({s^\otimes},a_*)-Q({s^\otimes},a_*)+\gamma \max_{a'}(Q(s^\otimes_*,a'))]$\;
				$ \bar{P}^\otimes(s^\otimes_i,a,s^\otimes_j)\leftarrow \psi(s^\otimes_i,a,s^\otimes_j)/\Psi(s^\otimes_i,a),~\forall s^\otimes_i,s^\otimes_j\in \mathcal{S}^\otimes ~\text{and}~ \forall a \in \mathcal{A}$\;
				$\mathit{PSP}(s^\otimes)\leftarrow \max_{a \in \mathcal{A}}\sum_{{s^\otimes}' \in \mathcal{S}^\otimes} \bar{P}^\otimes(s^\otimes,a,{s^\otimes}') \times \mathit{PSP}({s^\otimes}')$\;
				$s^\otimes=s^\otimes_*$\;
			}
		}
	\end{small}
	\caption{Logically-Constrained RL}
	\label{algorithm}
\end{algorithm2e}

In the standard value iteration method the value estimation is simultaneously updated for all states. However, an alternative method is to update the value for one state at a time. This method is known as asynchronous value iteration. 

\begin{definition}
	[Gauss-Seidel Asynchronous Value Iteration (AVI)\cite{NDP}]
	We denote AVI operation by $F$ and is defined as follows:
	
	\begin{equation}
	F~\mathit{PSP}(s_1)=\max\limits_{a \in \mathcal{A}}\{\sum_{s' \in \mathcal{S}} P(s_1,a,s') \mathit{PSP}(s')\},
	\end{equation}
	where $ s_1 $ is the state that current state at the MDP, and for all $s_i \neq s_1$:
	
	\begin{align}\label{gauss}
	\begin{aligned}
	F~\mathit{PSP}(s_i)=&\max\limits_{a \in \mathcal{A}}\{\sum_{s' \in \{s_1,...,s_{i-1}\}} P(s_i,a,s')~\times\\
	F~\mathit{PSP}(s')+&\sum_{s' \in \{s_i,...,s_{|\mathcal{S}|}\}} P(s_i,a,s') \mathit{PSP}(s')\}.
	\end{aligned}
	\end{align} \hfill$\lrcorner$
\end{definition} 

By (\ref{gauss}) we update the value of $\mathit{PSP}$ state by state and use the calculated value for the next step.

\begin{proposition}[From \cite{NDP}]\label{AVI}
	Let $k_0, k_1, ...$ be an increasing sequence of iteration indices such that $k_0=0$ and each state is updated at least once between iterations $k_m$ and $k_{m+1}-1$, for all $m=0,1,...$. Then the sequence of value vectors generated by AVI asymptotically converges to $\underline{\mathit{PSP}}^*$. \hfill$\lrcorner$ 
\end{proposition}

\begin{lemma}[From \cite{NDP}]\label{contraction}
	$F$ is a contraction mapping with respect to the infinity norm. In other words, for any two value vectors $\mathit{PSP}$ and $\mathit{PSP}'$:
	
	\begin{equation*}
	||F~\underline{\mathit{PSP}}-F~\underline{\mathit{PSP}}'||_\infty \leq ||\underline{\mathit{PSP}}-\underline{\mathit{PSP}}'||_\infty.
	\end{equation*}
	\hfill$\lrcorner$
\end{lemma}

\begin{proposition}
	[Convergence, \cite{alphaMDP}] From Lemma \ref{contraction}, and under the assumptions of Proposition \ref{AVI}, $ \bar{P}(s,a,s') $ converges to $ P(s,a,s') $, and from Proposition \ref{AVI}, 
	the AVI value vector $ \underline{\mathit{PSP}} $ asymptotically converges to $\underline{\mathit{PSP}}^*$, i.e. the probability that could be alternatively calculated by DP-based methods if the MDP was fully known.
	\hfill$\lrcorner$
\end{proposition}

\noindent We conclude this section by presenting the overall procedure in Algorithm \ref{algorithm}. 
The input of LCRL includes $ \textit{it\_threshold}$, which is an upper bound on the number of iterations. 

\section{Experimental Results}
\label{case study}
We discuss a number of safety-critical motion planning experiments, namely policy synthesis problems around temporal specifications that are extended with safety requirements. 

\begin{figure}[!t] \centering \includegraphics[width=0.8\textwidth]{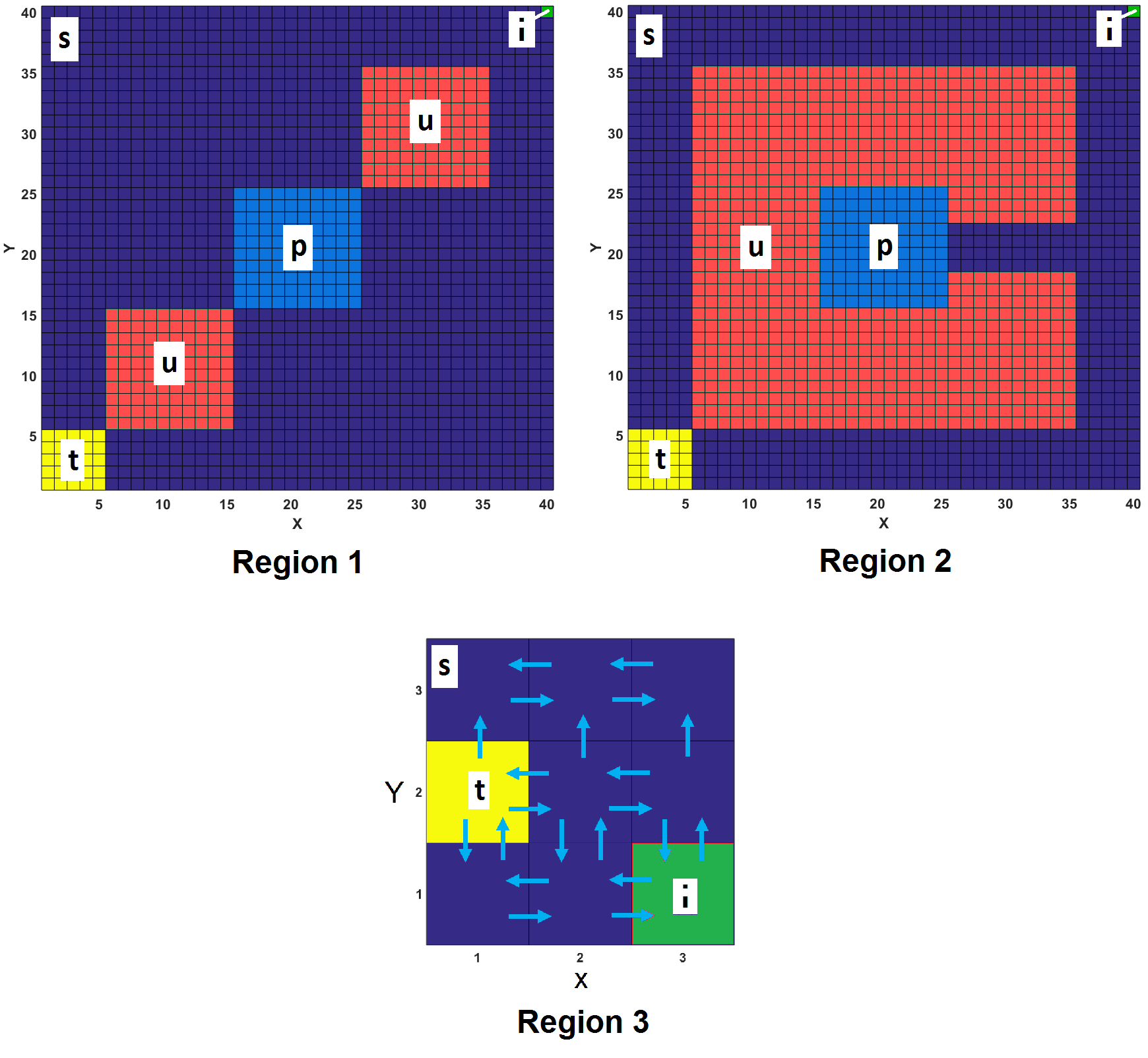} \caption{MDP labelling~-~green: initial state ($i$), dark blue: safe ($s$), red: unsafe ($u$), light blue: pre-target ($p$), yellow: target ($t$)}
	\label{MDPs} 
\end{figure}
\begin{figure}[!t]
	\centering
	\includegraphics[width=0.6\textwidth]{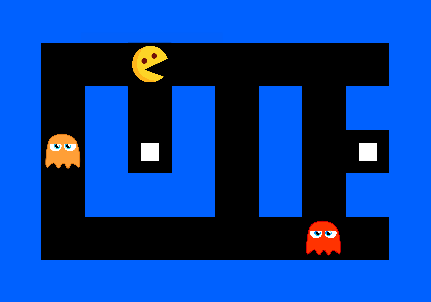}
	\caption{Pacman environment (initial state) - dot on the left is labeled as food 1 ($ f_1 $) and the one on the right as food 2 ($ f_2 $), the state of crashing to a ghost is labeled as $ g $ and the rest is neutral ($ n $).}
	\label{pacmaninit} 
\end{figure}

\subsection{Description of the Benchmark}

\textbf{The first experiment} is an LTL-constrained control synthesis problem for a robot in a grid-world. Let the grid be an $ L \times L $ square over which the robot moves. In this setup, the robot location is the MDP state $s \in \mathcal{S} $. At each state $s \in \mathcal{S}$ the robot has a set of actions $\mathcal{A}=\{\mathit{left},\mathit{right},\mathit{up},\mathit{down},\mathit{stay}\}$ by which the robot is able to move to other states (e.g. $s'$) with the probability of $P(s,a,s'), a \in \mathcal{A}$. At each state $s \in \mathcal{S}$, the actions available to the robot are either to move to a neighbor state $s' \in \mathcal{S}$ or to stay at the state $s$. In this example, if not otherwise specified, we assume for each action the robot chooses, there is a probability of $85\%$ that the action takes the robot to the correct state and $15\%$  that the action takes the robot to a random state in its neighbourhood (including its current state). This example is a well-known benchmark in the machine leaning community and is often referred to as ``slippery grid-world''.

A labelling function $L:\mathcal{S}\rightarrow 2^{\mathcal{AP}}$ assigns to each state $s \in \mathcal{S}$ a set of atomic propositions $L(s) \subseteq \mathcal{AP}$. We assume that in each state $s$ the robot is aware of the labels of the neighbouring states, which is a realistic assumption. We consider two  $40 \times 40$ regions and one $3 \times 3$ region with different labellings as in Fig. \ref{MDPs}. In Region 3 and in state target, the subsequent state after performing action $stay$ is always the state target itself. Note that all the actions are not active in Region 3 and the agent has to avoid the top row otherwise it gets trapped.

\textbf{The second experiment} is a version of the well-known Atari game Pacman (Fig. \ref{pacmaninit}) that is initialized in a tricky configuration. In order to win the game the agent has to collect all available tokens without being caught by moving ghosts. The ghosts' movement is stochastic and there is a probability $ p_g $ for each ghost that determines if the ghost is chasing Pacman (often referred to as ``chase mode'') or if it is executing a random action (``scatter mode''). Notice that unlike the first experiment, in this setup the actions of the ghosts and of the agent result in a deterministic transition, i.e. the world is not ``slippery''.  

\subsection{Properties}

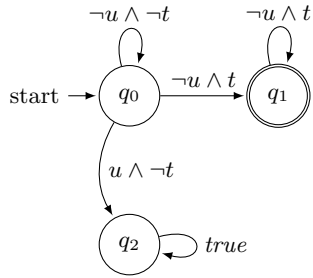
\begin{figure}[!t]\centering
	\scalebox{.99}{
		\begin{tikzpicture}[shorten >=1pt,node distance=2cm,on grid,auto] 
		\node[state,initial] (q_0)   {$q_0$}; 
		\node[state,accepting] (q_1) [right=of q_0] {$q_1$}; 
		\node[state] (q_2) [below=of q_0] {$q_2$}; 
		\path[->] 
		(q_0) edge  node {$\neg u \wedge t$} (q_1)
		(q_0) edge [bend right] node {$u \wedge \neg t$} (q_2)
		edge [loop above] node {$\neg u \wedge \neg t$} ()
		(q_1) edge [loop above] node {$\neg u \wedge t$} ()
		(q_2) edge [loop right] node {$\mathit{true}$} ();
		\end{tikzpicture}}
	\caption{LDBA for (\ref{ltl}) with removed transitions labelled $t \wedge u$ (since it is impossible to be at target and unsafe at the same time)}
	\label{f4}  
\end{figure}

\begin{figure}[!t]\centering
	\scalebox{.99}{
		\begin{tikzpicture}[shorten >=1pt,node distance=2cm,on grid,auto] 
		\node[state,initial] (q_0)   {$q_0$}; 
		\node[state,accepting] (q_1) [right=of q_0] {$q_1$}; 
		\path[->] 
		(q_0) edge  node {$\varepsilon$} (q_1)
		edge [loop below] node {$\mathit{true}$} ()
		(q_1) edge [loop below] node {$t$} ();
		\end{tikzpicture}}
	\caption{LDBA for (\ref{ltl2})}
	\label{f5} 
\end{figure}

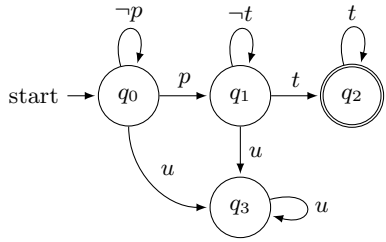
\begin{figure}[!t]\centering
	\scalebox{.99}{
		\begin{tikzpicture}[shorten >=1pt,node distance=1.5cm,on grid,auto] 
		\node[state,initial] (q_1)   {$q_0$}; 
		\node[state] (q_2) [right=of q_1] {$q_1$}; 
		\node[state] (q_5) [below=of q_2] {$q_3$}; 
		\node[state,accepting] (q_3) [right=of q_2] {$q_2$}; 
		\path[->] 
		(q_1) edge [loop above] node {$\neg p$} ()   	
		(q_1) edge  node {$p$} (q_2)
		(q_2) edge [loop above] node {$\neg t$} ()
		(q_2) edge node {$t$} (q_3)
		(q_3) edge [loop above] node {$t$} () 
		(q_1) edge [bend right=45] node {$u$} (q_5)
		(q_2) edge node {$u$} (q_5)
		(q_5) edge [loop right] node {$u$} ();
		\end{tikzpicture}}
	\caption{LDBA for (\ref{omega})}
	\label{f3}
\end{figure} 

\textbf{In the first experiment (slippery grid-world)}, we consider the following LTL properties. The two first properties \eqref{ltl} and \eqref{ltl2} focus on safety and reachability while the third property \eqref{omega} requires a sequential visit to states with label $ p $ and then target $ t $.

\begin{equation}
\label{ltl}
\lozenge t  \wedge  \square(t \rightarrow \square t) \wedge \square(u \rightarrow \square u),
\end{equation}
\begin{equation}
\label{ltl2}
\lozenge \square t,
\end{equation}
and
\begin{equation}
\label{omega}
\lozenge(p \wedge \lozenge t) \wedge \square(t \rightarrow \square t) \wedge \square(u \rightarrow \square u),  
\end{equation}
where $t$ stands for ``target'', $u$ stands for ``unsafe'', and $p$ refers to the area that has to be visited before visiting the area with label $t$. The property \eqref{ltl} asks the agent to eventually find the target $ \lozenge t $ and stays there $ \square(t \rightarrow \square t) $ while avoiding the unsafe otherwise it is going to be trapped there $ \square(u \rightarrow \square u) $. The specification \eqref{ltl2} requires the agent to eventually find the target and stays there and the intuition behind \eqref{omega} is that the agent has to eventually first visit $ p $ and then visit $ t $ at some point in the future $ \lozenge(p \wedge \lozenge t) $ and stays there $ \square(t \rightarrow \square t) $ while avoiding unsafe areas $ \square(u \rightarrow \square u) $. 

We can build the B\"uchi automaton associated with
(\ref{ltl}), (\ref{ltl2}) and (\ref{omega}) as in Fig.~\ref{f4}, Fig.~\ref{f5} and Fig.~\ref{f3}. Note that the LDBA in Fig.~\ref{f4} has a non-deterministic $\varepsilon$ transition.

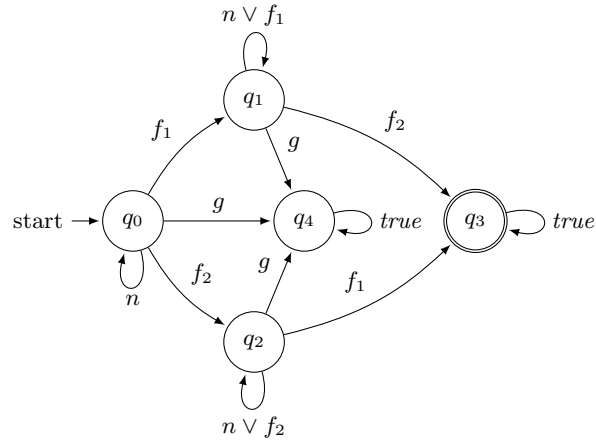
\begin{figure}[!t]\centering
	\scalebox{.99}{
		\begin{tikzpicture}[shorten >=1pt,node distance=2.3cm,on grid,auto] 
		\node[state,initial] (q_1) {$q_0$}; 
		\node[state] (q_2) [above right=of q_1] {$q_1$}; 
		\node[state] (q_3) [below right=of q_1] {$q_2$}; 
		\node[state] (q_5) [right=of q_1] {$q_4$}; 
		\node[state,accepting] (q_4) [right=of q_5] {$q_3$}; 
		\path[->] 
		(q_1) edge [loop below] node {$n$} ()   	
		(q_1) edge [bend left=15] node {$f_1$} (q_2)
		(q_1) edge [bend right=15] node {$f_2$} (q_3)
		(q_2) edge [loop above] node {$n \vee f_1$} ()
		(q_2) edge [bend right=-15] node {$f_2$} (q_4)
		(q_3) edge [loop below] node {$n \vee f_2$} ()
		(q_3) edge [bend right=15] node {$f_1$} (q_4) 
		(q_2) edge node {$g$} (q_5)
		(q_3) edge node {$g$} (q_5)
		(q_1) edge node {$g$} (q_5)
		(q_5) edge [loop right] node {$\mathit{true}$} ()
		(q_4) edge [loop right] node {$\mathit{true}$} ();
		\end{tikzpicture}}
	\caption{LDBA for (\ref{pacman_p})}
	\label{pacman_a}
\end{figure} 

\textbf{In the second experiment (Pacman)}, to win the game, Pacman is required to choose between one of the two available foods and then find the other one ($ \lozenge [ (f_1 \wedge \lozenge f_2) \vee (f_2 \wedge \lozenge f_1)] $) while avoiding ghosts ($ \square(g \rightarrow \square g) $). Thus, the property of interest in this game is
\begin{equation}
\label{pacman_p}
\lozenge [ (f_1 \wedge \lozenge f_2) \vee (f_2 \wedge \lozenge f_1)]  \wedge \square(g \rightarrow \square g),  
\end{equation}
The associated B\"uchi automaton is shown in Fig. \ref{pacman_a}. 

\subsection{Simulation Results}

\begin{figure}[!t]
	\centering
	\includegraphics[width=0.8\textwidth]{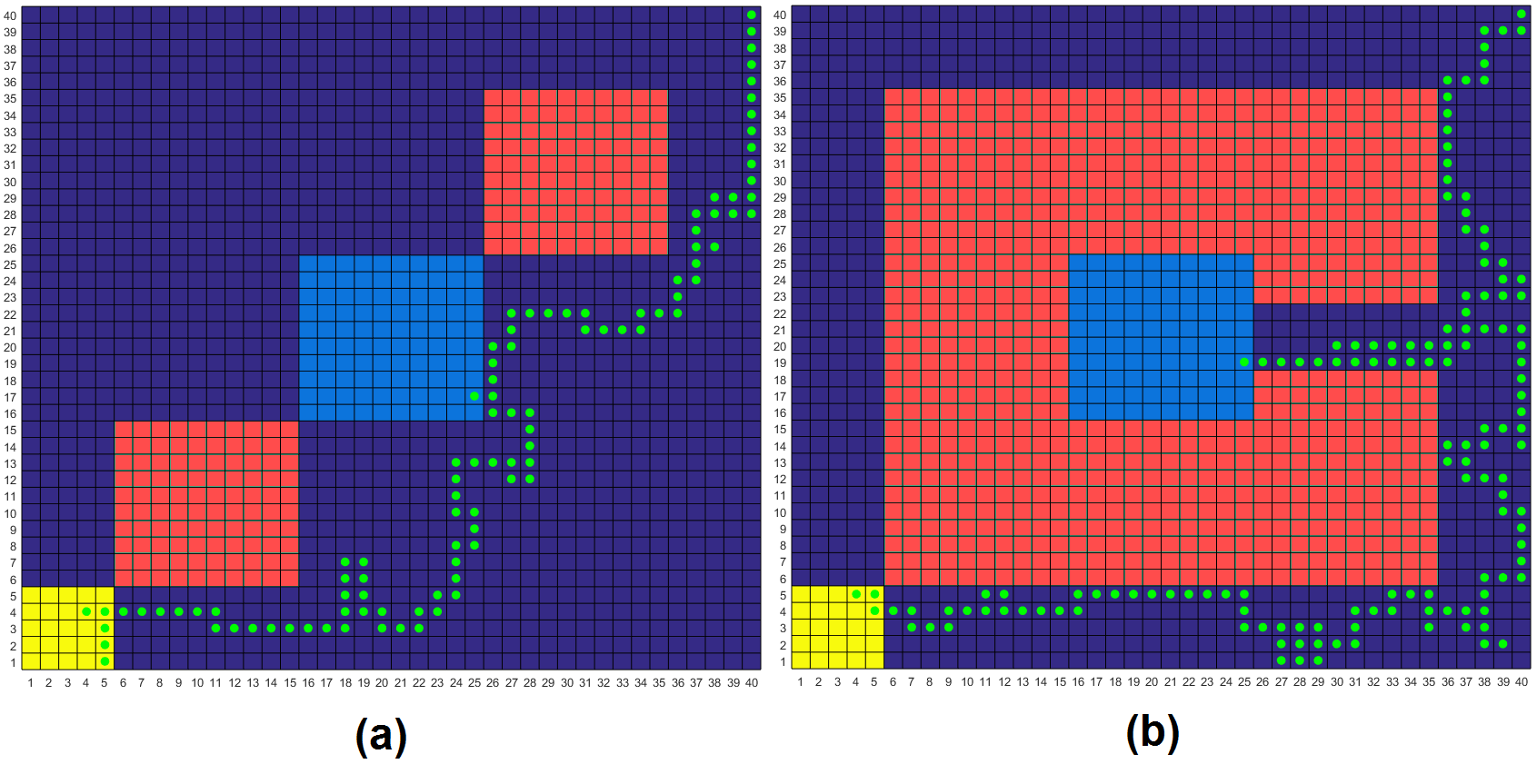}
	\caption{Simulation results for (\ref{omega}) in (a) Region~1 and (b) Region~2}
	\label{MDP1and2omega} 
\end{figure}

\begin{figure}[!t]
	\centering
	\includegraphics[width=0.8\textwidth]{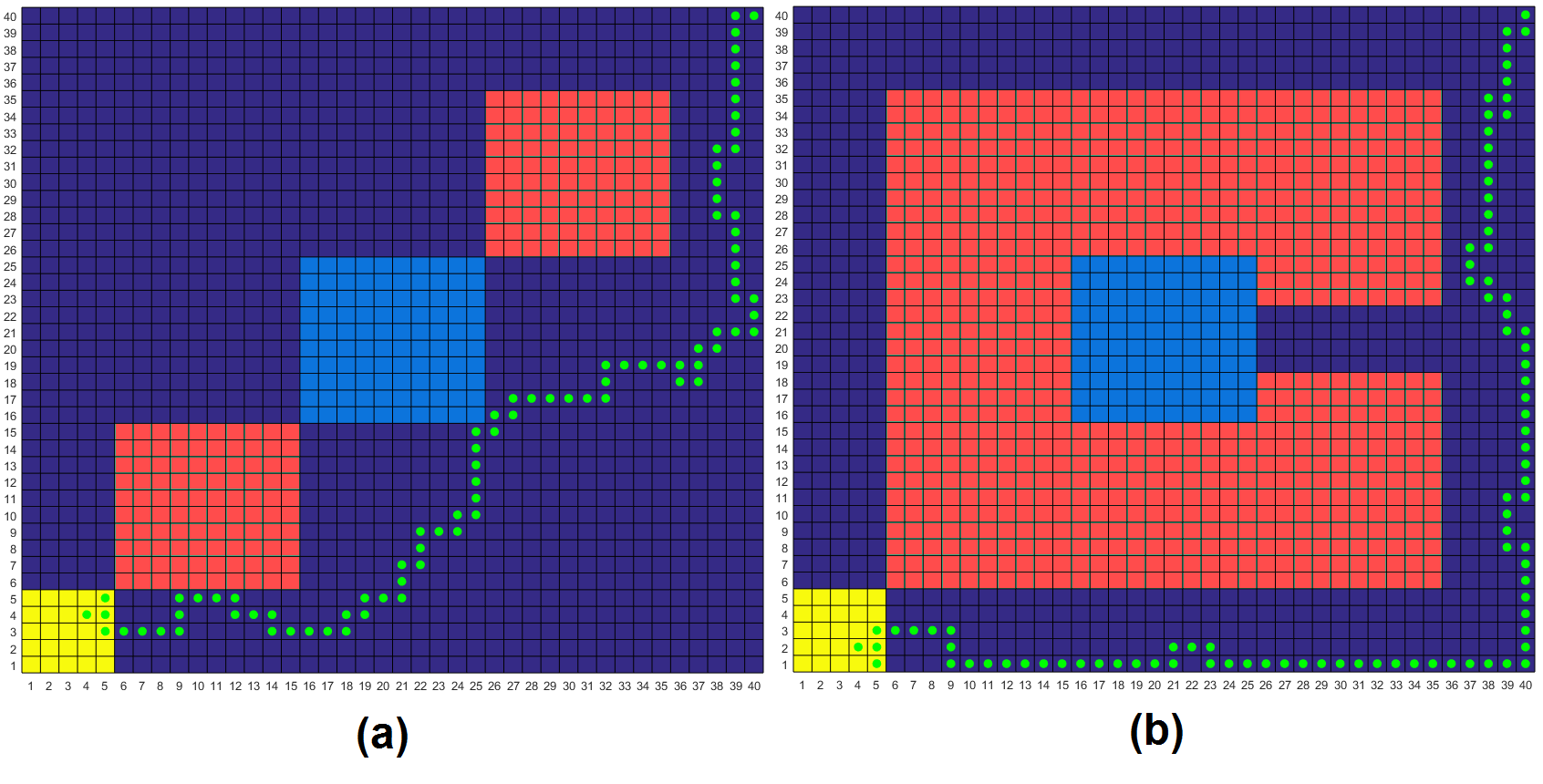}
	\caption{Simulation results for (\ref{ltl}) in (a) Region 1 and (b) Region 2}
	\label{MDP1and2ltl} 
\end{figure}

\begin{figure}[!t]
	\centering
	\includegraphics[width=0.25\textwidth]{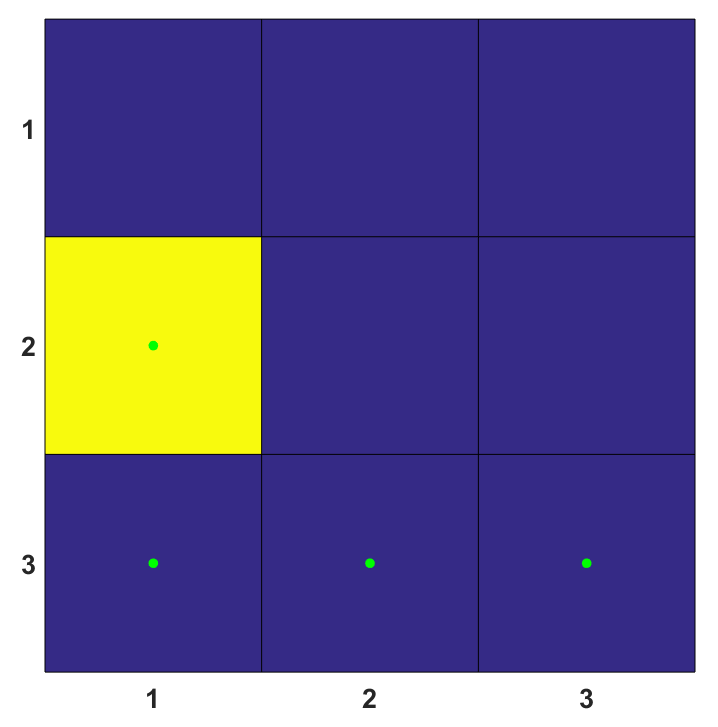}
	\caption{Simulation results for (\ref{ltl2}) in Region~3}
	\label{MDP3ltl} 
\end{figure}

\begin{figure}[!t]
\centering
\includegraphics[width=0.9\textwidth]{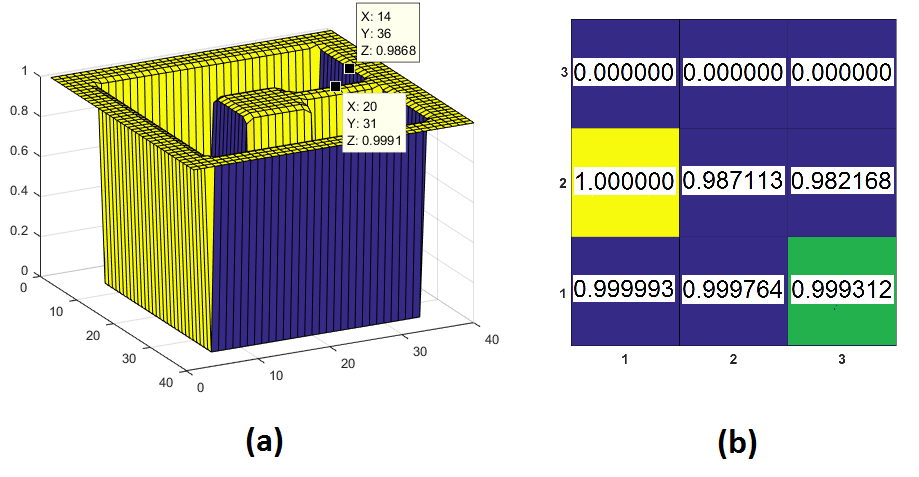}
\caption{Maximum probability of satisfying the LTL property in
(a) Region~2 and (b) Region~3: The result generated by our proposed algorithm is identical to PRISM result}
\label{PRISM1} 
\end{figure}


\begin{figure}[!t]
\centering
\includegraphics[width=0.6\textwidth]{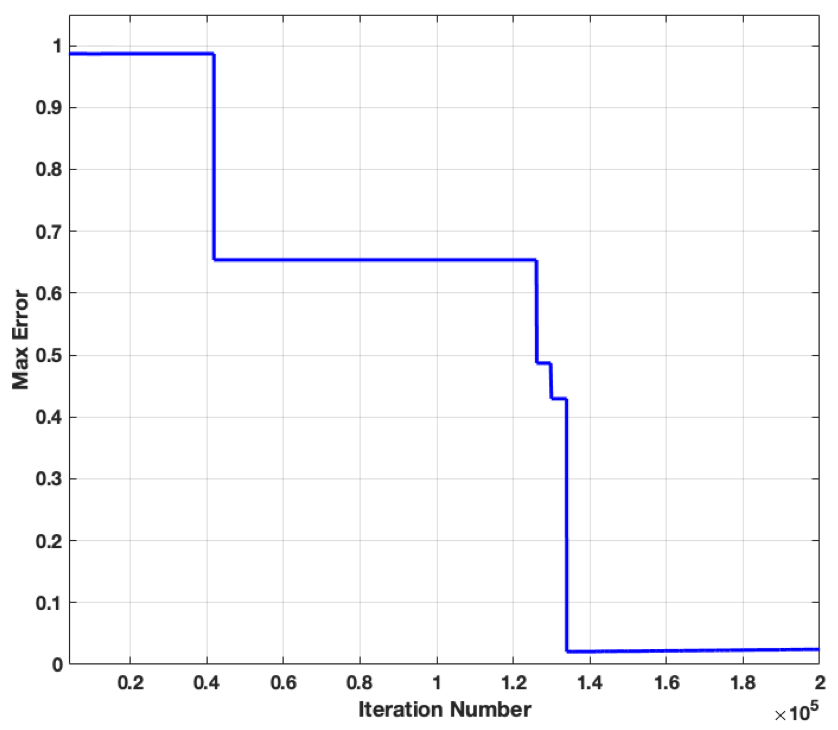}
\caption{Maximum error between the formula likelihood computed with PRISM and with our algorithm}
\label{max_error} 
\end{figure}

\begin{figure}[!t]
\centering
\includegraphics[width=0.6\textwidth]{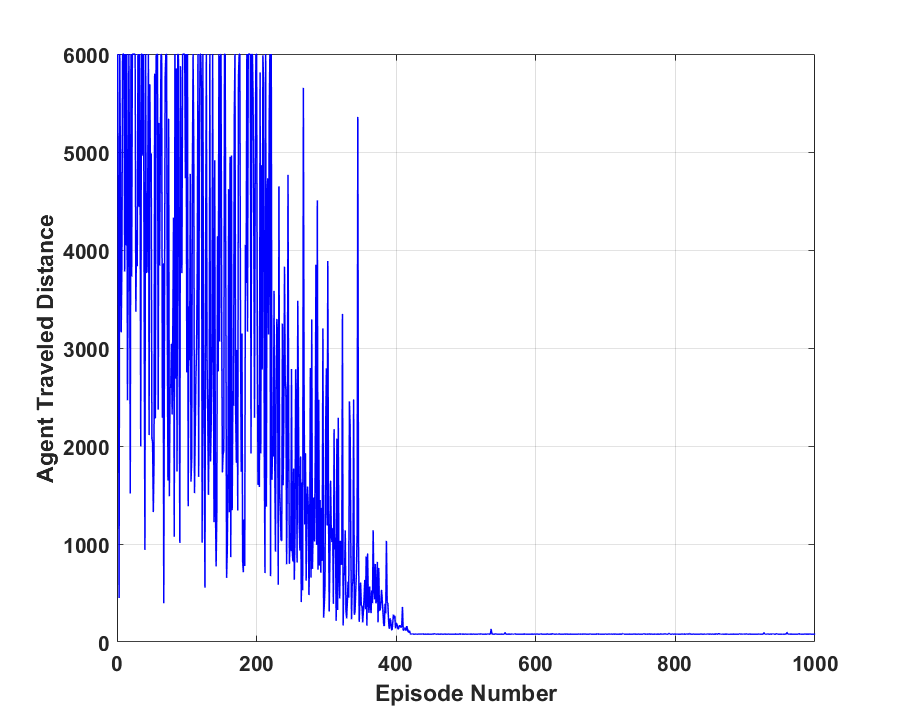}
\caption{The distance that agent traverses from initial state to final state in Region 1 under (\ref{omega})}
\label{converging} 
\end{figure}

\textbf{In the first experiment (slippery grid-world)}, the simulation parameters are set as $\mu=0.9$ and $\gamma=0.9$. Fig.~\ref{MDP1and2omega} gives the results of the
learning for the expression (\ref{omega}) in Region 1 and Region 2 after
$400,000$ iterations and $200$ episodes. According to (\ref{omega}) the robot
has to avoid the red (unsafe) regions, visit the light-blue (pre-target)
area at least once and then go to the yellow (target) region. Recall that
the robot desired action is executed with a probability of $85\%$. Thus,
there might be some undesired deviations in the robot path.

Fig.~\ref{MDP1and2ltl} gives the results of the learning
for the LTL formula~(\ref{ltl}) in Region 1 and Region 2 after $400,000$
iterations and $200$ episodes. The intuition behind the LTL formulae
in~(\ref{ltl}) is that the robot has to avoid red (unsafe) areas until it
reaches the yellow (target) region, otherwise the robot is going to be stuck
in the red (unsafe) area.

Finally, in Fig.~\ref{MDP3ltl} the learner tries to satisfy the LTL formula
$\lozenge \square t$ in~(\ref{ltl2}). The learning takes 1000 iterations and
20 episodes.

Fig.~\ref{PRISM1} gives the result of our proposed
value iteration method for calculating the maximum probability
of satisfying the LTL property in Region 2 and Region 3. In both cases our
method was able to accurately calculate the maximum probability of
satisfying the LTL property. We observed convergence to zero in the maximum error between the correct $\mathit{PSP}^*$ calculated by PRISM and the probability calculation by our proposed algorithm (Fig. \ref{max_error}.a). 

Fig. \ref{max_error}.b shows the distance that agent traverses from initial state to final state at each episode in Region 1 under (\ref{omega}). After almost 400 episodes of learning the agent converges to the final optimal policy and the travelled distance stabilizes. 

\begin{figure}[!t] \centering \includegraphics[width=\textwidth]{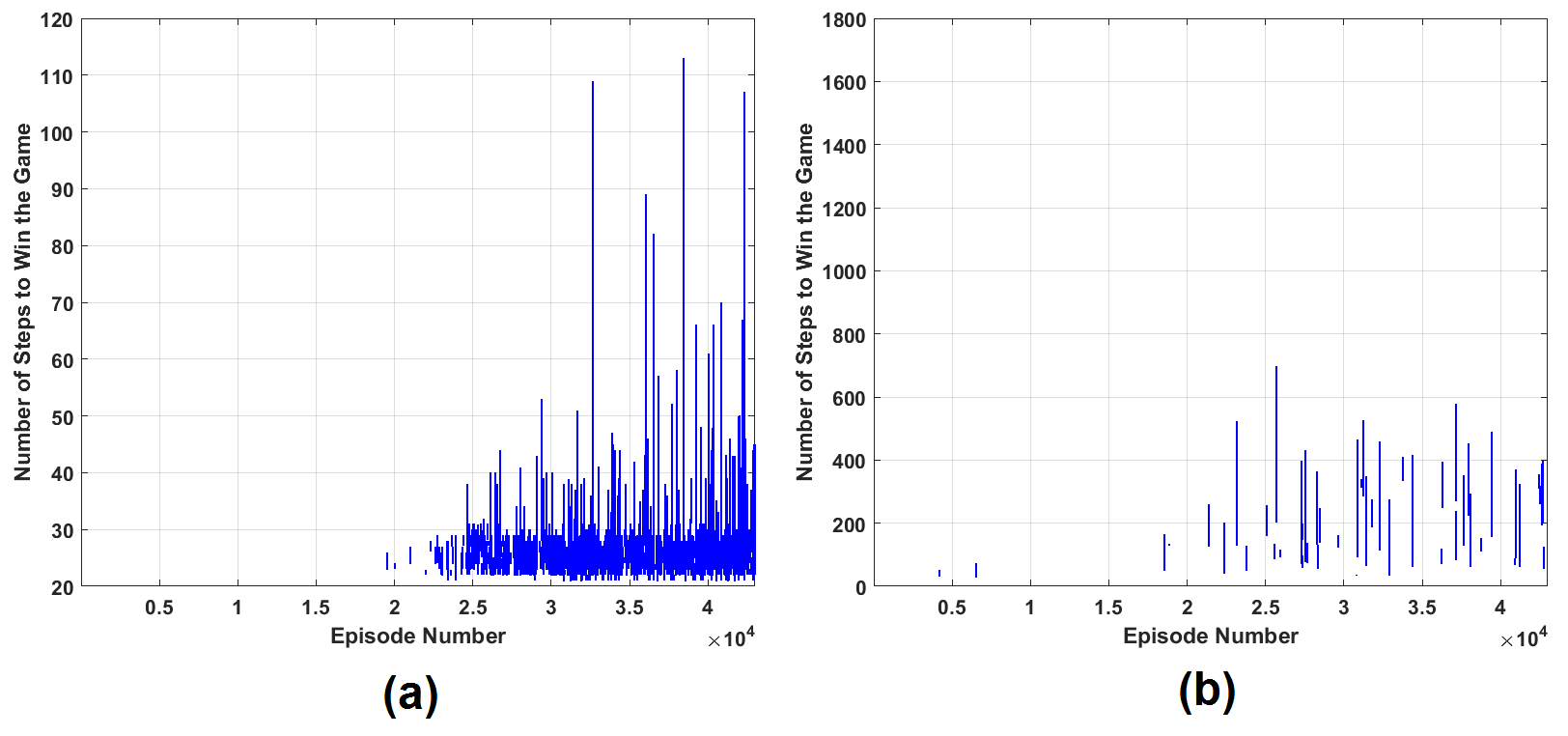} \caption{Results$^*$ of learning in Pacman with (a) LCRL (b) classical QL with positive reward for winning the game \\ ${}^*$ blank space means Pacman lost the game} 
	\label{pacman_results} 
\end{figure}

\textbf{In the second experiment (Pacman)}, the simulation parameters are set as $\mu=0.9$ and $\gamma=0.9$. The stochastic behaviour of the ghosts is also captured by $ p_g=0.9 $. Fig.~\ref{pacman_results} gives the results of learning with LCRL\footnote{please visit \url{www.cs.ox.ac.uk/conferences/lcrl/} to watch the videos of the agent playing Pacman. The code is adaptive and you can try your own configuration as well!} and classical RL for (\ref{pacman_a}). After almost 20,000 episodes, LCRL finds a stable policy to win the game even with ghosts playing probabilistically. The average steps for the agent to win the game (y axis) in LCRL is around 25 steps which is very close to human-level performance of 23 steps if the ghosts act deterministically. On the other hand, standard RL (in this case, classical QL with positive reward for winning the game) fails to find a stable policy and only achieved a number of random winnings with huge numbers of steps.  

\subsection{Comparison with a DRA-based algorithm}
\begin{figure}[!t] \centering \includegraphics[width=0.7\textwidth]{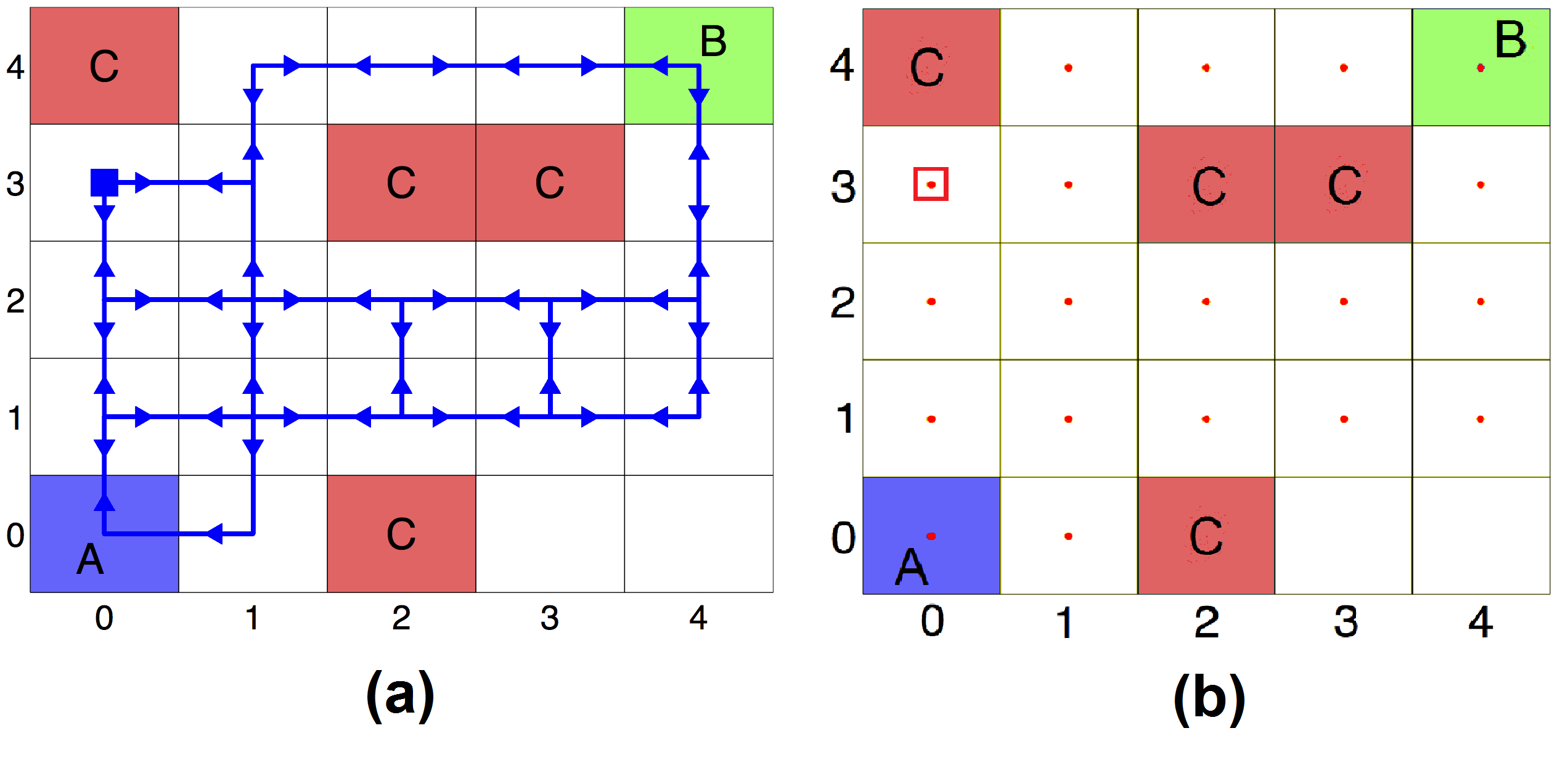} \caption{(a) Example considered in \cite{dorsa} (b) Trajectories under the policy generated by LCRL in \cite{dorsa} example}
\label{dorsaandus} 
\end{figure}

The problem of LTL-constrained learning is also investigated in~\cite{dorsa}, where the
authors propose to translate the LTL property into a DRA and then to construct a product MDP. A $5\times 5$ grid world is considered and starting from state $(0,3)$ the agent has to visit two regions infinitely often (areas $A$ and $B$ in
Fig.~\ref{dorsaandus}). The agent has to also avoid the area~$C$. 
This property can be encoded as the following LTL formula: 
\begin{equation}
\label{dorsa's ltl}
\square\lozenge A \wedge \square\lozenge B \wedge \square \neg C.
\end{equation} 

\begin{figure}[!t]\centering
\scalebox{.8}{
\begin{tikzpicture}[shorten >=1pt,node distance=2.8cm,on grid,auto] 
   \node[state,initial,accepting] (q_0)   {$q_0$}; 
   \node[state] (q_1) [above right=of q_0] {$q_1$}; 
   \node[state] (q_2) [below right=of q_1] {$q_2$}; 
   \path[->] 
    (q_0) edge  node {$\neg B \wedge \neg C$} (q_1)
    (q_0) edge node {$B \wedge \neg A \wedge \neg C$} (q_2)
    (q_2) edge [bend left=45] node {$A \wedge \neg C$} (q_0)
    (q_1) edge  node {$B \wedge \neg A \wedge \neg C$} (q_2)
    (q_2) edge [loop right] node {$\neg A \wedge \neg C$} ()
    (q_1) edge [loop above] node {$\neg B \wedge \neg C$} ();    
\end{tikzpicture}}
\caption{LDBA expressing the LTL formula in (\ref{dorsa's ltl}) with removed transitions labelled $A \wedge B$ (since it is impossible to be at $A$ and $B$ at the same time)}
\label{dorsa3} 
\end{figure}
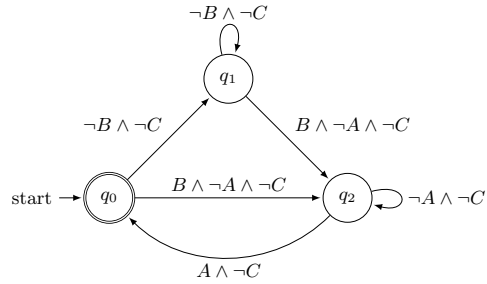

The product MDP in~\cite{dorsa} contains 150 states, which means that the
Rabin automaton has 6 states. Fig.~\ref{dorsaandus}.a shows the trajectories under the optimal policy generated by \cite{dorsa} algorithm after 600 iterations. However, by employing LCRL we are able to generate the same trajectories with only 50 iterations (Fig.~\ref{dorsaandus}.b). The automaton that we consider is an LDBA with only 3 states as in Fig.~\ref{dorsa3}. This result in a smaller product MDP and a much more succinct state space (only 75 states) for the algorithm to learn, which consequently leads to a faster convergence. 

In addition, the reward shaping in LCRL is significantly simpler thanks to the B\"uchi acceptance condition. In a DRA $\textbf{R}=(\mathcal{Q},\mathcal{Q}_0,\Sigma, \mathcal{F},
\Delta)$, the set $\mathcal{F}= \{(G_1,B_1),\ldots,\allowbreak
(G_{n_F},\allowbreak B_{n_F})\}$ represents the acceptance condition in which $\allowbreak G_i,\allowbreak B_i \in \mathcal{Q}$ for $i=1,\ldots,\allowbreak n_F$. An infinite run $\theta \in
\mathcal{Q}^\omega$ starting from $\mathcal{Q}_0$ is accepting if there
exists $i\in \{1,\ldots,n_F\}$ such that
$$
\mathit{inf}(\theta) \cap {G}_i \neq  \emptyset
\quad\mbox{and}\quad \mathit{inf}(\theta) \cap {B}_i =  \emptyset.
$$ 
Therefore for each $i\in \{1,\ldots,\allowbreak n_F\}$ a separate reward assignment is
needed in~\cite{dorsa} which complicates the implementation and increases the required calculation costs. More importantly, LCRL is a model-free learning algorithm that does not require an approximation of the transition probabilities of the underlying MDP. This even makes LCRL more easier to employ. We would like to emphasize that LCRL convergence proof solely depends on the structure of the MDP and this allows LCRL to find satisfying policies even if they have probability of less than one.

\section{Conclusion}

In this paper we have proposed a method to constrain RL resulting traces by
using an LTL property. The proposed algorithm is the first work on constraining model-free RL by a temporal logic specification which can be a foundation for the future work on this direction. We have argued that converting the LTL property to an LDBA results in a significantly smaller automaton than to a DRA (the standard approach in the literature), which decreases the size of the product MDP and increases RL convergence rate \cite{sickert}. 

A DRA needs more complicated accepting conditions than a B\"uchi automaton; thus,
a more complex reward assignment is required. Therefore, in addition to the
more succinct product MDP and faster convergence, our algorithm is easier to
implement as opposed to standard methods that convert the LTL property to a DRA. 

Additionally, we have proposed a value iteration method to calculate the probability of satisfaction of the LTL property. We argue that with this trajectory-based method we are able to extend the satisfaction probability calculation to large-scale MDPs which are hard for model-based methods (such as DP) to handle. The use of RL for policy generation allows the value iteration algorithm to focus on parts of state space that are relevant to the property. This results in a faster calculation of probability values when compared to DP, where these values are updated for the whole state space. The efficiency of DP is hindered by excessive memory requirements, caused by the need to store a full-model in memory and to apply the Bellman operator over the entire state-space. Thus, the main reason that LCRL improves performance and scalability is by avoiding an exhaustive update over the whole state space. In other words, LCRL “guides" the MDP exploration so as to minimize the solution time by only considering the portion of the MDP that is relevant to the LTL property. This is in particular made clear in the Pacman experiment, where we show that classical RL is very slow and ineffective in finding the optimal policy, whereas LCRL converges extremely fast thanks to the guidance that is provided by the automaton. 

The most immediate direction for future research is to extend this framework to continuous-state or continuous-action MDPs and investigate the theoretical guarantees that can be derived. It is also interesting to explore partially observable MDPs in which the agent is not sure about its current state and still has to take the optimal action.

\clearpage
\bibliographystyle{splncs}
\bibliography{Biblio}

\begin{thebibliography}{10}

\bibitem{puterman}
Puterman, M.L.:
\newblock {M}arkov decision processes: {D}iscrete stochastic dynamic
  programming.
\newblock John Wiley \& Sons (2014)

\bibitem{sutton2}
Sutton, R.S., Precup, D., Singh, S.:
\newblock Between {MDPs} and semi-{MDPs}: A framework for temporal abstraction
  in reinforcement learning.
\newblock Artificial intelligence \textbf{112}(1-2) (1999)  181--211

\bibitem{rahili}
Rahili, S., Ren, W.:
\newblock Game theory control solution for sensor coverage problem in unknown
  environment.
\newblock In: CDC, IEEE (2014)  1173--1178

\bibitem{ifac}
Hasanbeig, M., Pavel, L.:
\newblock On synchronous binary log-linear learning and second order
  {Q}-learning.
\newblock In: The 20th World Congress of the International Federation of
  Automatic Control (IFAC), IFAC (2017)

\bibitem{thesis}
Hasanbeig, M.:
\newblock Multi-agent learning in coverage control games.
\newblock Master's thesis, University of Toronto (Canada) (2016)

\bibitem{dorsa}
Sadigh, D., Kim, E.S., Coogan, S., Sastry, S.S., Seshia, S.A.:
\newblock A learning based approach to control synthesis of {Markov} decision
  processes for linear temporal logic specifications.
\newblock In: CDC, IEEE (2014)  1091--1096

\bibitem{ng}
Abbeel, P., Coates, A., Quigley, M., Ng, A.Y.:
\newblock An application of reinforcement learning to aerobatic helicopter
  flight.
\newblock NIPS \textbf{19} (2007) ~1

\bibitem{mnih}
Mnih, V., Kavukcuoglu, K., Silver, D., Rusu, A.A., Veness, J., Bellemare, M.G.,
  Graves, A., Riedmiller, M., Fidjeland, A.K., Ostrovski, G.,  et~al.:
\newblock Human-level control through deep reinforcement learning.
\newblock Nature \textbf{518}(7540) (2015)  529--533

\bibitem{silver}
Silver, D., Huang, A., Maddison, C.J., Guez, A., Sifre, L., Van Den~Driessche,
  G., Schrittwieser, J., Antonoglou, I., Panneershelvam, V., Lanctot, M.,
  et~al.:
\newblock Mastering the game of {G}o with deep neural networks and tree search.
\newblock Nature \textbf{529}(7587) (2016)  484--489

\bibitem{otterlo}
van Otterlo, M., Wiering, M.:
\newblock Reinforcement learning and {M}arkov decision processes.
\newblock In: Reinforcement Learning.
\newblock Springer (2012)  3--42

\bibitem{pnueli}
Pnueli, A.:
\newblock The temporal logic of programs.
\newblock In: Foundations of Computer Science, IEEE (1977)  46--57

\bibitem{sutton}
Sutton, R.S., Barto, A.G.:
\newblock Reinforcement learning: An introduction. Volume~1.
\newblock MIT press Cambridge (1998)

\bibitem{smith}
Smith, S.L., Tumov{\'a}, J., Belta, C., Rus, D.:
\newblock Optimal path planning for surveillance with temporal-logic
  constraints.
\newblock The International Journal of Robotics Research \textbf{30}(14) (2011)
   1695--1708

\bibitem{bible}
Baier, C., Katoen, J.P., Larsen, K.G.:
\newblock Principles of Model Checking.
\newblock MIT press (2008)

\bibitem{safra}
Safra, S.:
\newblock On the complexity of omega-automata.
\newblock In: Foundations of Computer Science, 1988., 29th Annual Symposium on,
  IEEE (1988)  319--327

\bibitem{nba2d}
Piterman, N.:
\newblock From nondeterministic {B{\"u}chi} and {Streett} automata to
  deterministic parity automata.
\newblock In: Logic in Computer Science, 2006 21st Annual IEEE Symposium on,
  IEEE (2006)  255--264

\bibitem{dra4}
Alur, R., La~Torre, S.:
\newblock Deterministic generators and games for {LTL} fragments.
\newblock TOCL \textbf{5}(1) (2004)  1--25

\bibitem{sickert}
Sickert, S., Esparza, J., Jaax, S., K{\v{r}}et{\'\i}nsk{\`y}, J.:
\newblock Limit-deterministic {B{\"u}chi} automata for linear temporal logic.
\newblock In: CAV, Springer (2016)  312--332

\bibitem{tkachev}
Tkachev, I., Mereacre, A., Katoen, J.P., Abate, A.:
\newblock Quantitative model-checking of controlled discrete-time {M}arkov
  processes.
\newblock Information and Computation \textbf{253} (2017)  1--35

\bibitem{wolf}
Wolff, E.M., Topcu, U., Murray, R.M.:
\newblock Robust control of uncertain {M}arkov decision processes with temporal
  logic specifications.
\newblock In: CDC, IEEE (2012)  3372--3379

\bibitem{topku}
Fu, J., Topcu, U.:
\newblock Probably approximately correct {MDP} learning and control with
  temporal logic constraints.
\newblock In: Robotics: Science and Systems X. (2014)

\bibitem{brazdil}
Br{\'a}zdil, T., Chatterjee, K., Chmel{\'\i}k, M., Forejt, V.,
  K{\v{r}}et{\'\i}nsk{\`y}, J., Kwiatkowska, M., Parker, D., Ujma, M.:
\newblock Verification of {M}arkov decision processes using learning
  algorithms.
\newblock In: ATVA, Springer (2014)  98--114

\bibitem{deepql}
Mnih, V., Kavukcuoglu, K., Silver, D., Rusu, A.A., Veness, J., Bellemare, M.G.,
  Graves, A., Riedmiller, M., Fidjeland, A.K., Ostrovski, G.,  et~al.:
\newblock Human-level control through deep reinforcement learning.
\newblock Nature \textbf{518}(7540) (2015)  529--533

\bibitem{asynchronous}
Mnih, V., Badia, A.P., Mirza, M., Graves, A., Lillicrap, T., Harley, T.,
  Silver, D., Kavukcuoglu, K.:
\newblock Asynchronous methods for deep reinforcement learning.
\newblock In: ICML. (2016)  1928--1937

\bibitem{belta2}
Svorenova, M., Cerna, I., Belta, C.:
\newblock Optimal control of {MDP}s with temporal logic constraints.
\newblock In: CDC, IEEE (2013)  3938--3943

\bibitem{game}
Wen, M., Ehlers, R., Topcu, U.:
\newblock Correct-by-synthesis reinforcement learning with temporal logic
  constraints.
\newblock In: IROS, IEEE (2015)  4983--4990

\bibitem{nils}
Junges, S., Jansen, N., Dehnert, C., Topcu, U., Katoen, J.P.:
\newblock Safety-constrained reinforcement learning for {MDP}s.
\newblock In: TACAS, Springer (2016)  130--146

\bibitem{belta}
Li, X., Vasile, C.I., Belta, C.:
\newblock Reinforcement learning with temporal logic rewards.
\newblock arXiv preprint arXiv:1612.03471 (2016)

\bibitem{scltl}
Kupferman, O., Vardi, M.Y.:
\newblock Model checking of safety properties.
\newblock Formal Methods in System Design \textbf{19}(3) (2001)  291--314

\bibitem{morteza}
Lahijanian, M., Wasniewski, J., Andersson, S.B., Belta, C.:
\newblock Motion planning and control from temporal logic specifications with
  probabilistic satisfaction guarantees.
\newblock In: ICRA, IEEE (2010)  3227--3232

\bibitem{pmc}
Pathak, S., Pulina, L., Tacchella, A.:
\newblock Verification and repair of control policies for safe reinforcement
  learning.
\newblock Applied Intelligence (2017)  1--23

\bibitem{andersson}
Andersson, O., Heintz, F., Doherty, P.:
\newblock Model-based reinforcement learning in continuous environments using
  real-time constrained optimization.
\newblock In: AAAI. (2015)  2497--2503

\bibitem{shield}
Alshiekh, M., Bloem, R., Ehlers, R., K{\"o}nighofer, B., Niekum, S., Topcu, U.:
\newblock Safe reinforcement learning via shielding.
\newblock arXiv preprint arXiv:1708.08611 (2017)

\bibitem{teacher}
Thomaz, A.L., Breazeal, C.:
\newblock Teachable robots: Understanding human teaching behavior to build more
  effective robot learners.
\newblock Artificial Intelligence \textbf{172}(6-7) (2008)  716--737

\bibitem{watkins}
Watkins, C.J., Dayan, P.:
\newblock {Q}-learning.
\newblock Machine learning \textbf{8}(3-4) (1992)  279--292

\bibitem{prism}
Kwiatkowska, M., Norman, G., Parker, D.:
\newblock {PRISM} 4.0: Verification of probabilistic real-time systems.
\newblock In: CAV, Springer (2011)  585--591

\bibitem{alphaMDP}
Kearns, M., Singh, S.:
\newblock Near-optimal reinforcement learning in polynomial time.
\newblock Machine learning \textbf{49}(2-3) (2002)  209--232

\bibitem{NDP}
Bertsekas, D.P., Tsitsiklis, J.N.:
\newblock Neuro-dynamic Programming. Volume~1.
\newblock Athena Scientific (1996)

\bibitem{stochastic}
Durrett, R.:
\newblock Essentials of stochastic processes. Volume~1.
\newblock Springer (1999)

\bibitem{pareto}
Forejt, V., Kwiatkowska, M., Parker, D.:
\newblock Pareto curves for probabilistic model checking.
\newblock In: ATVA, Springer (2012)  317--332

\bibitem{dis2undis}
Feinberg, E.A., Fei, J.:
\newblock An inequality for variances of the discounted rewards.
\newblock Journal of Applied Probability \textbf{46}(4) (2009)  1209--1212

\end{thebibliography}

\clearpage
\appendix
\section*{Appendix 1}
\textbf{Theorem \ref{thm}}. Let MDP $\textbf{M}_\textbf{N}$ be the product of an MDP~$\textbf{M}$ and an
automaton $\textbf{N}$ where $\textbf{N}$ is the LDBA associated with the
desired LTL property $\varphi$. If an accepting policy exists then LCRL optimal policy, which optimizes the expected utility, will find this policy. \\
\\
Proof:\\
\noindent
We would like to emphasise that the following proof holds for any GBA and it is not restricted to LDBAs. 

Assume that there exists a policy
$\overline{\mathit{Pol}}$ that satisfies $\varphi$. Policy $\overline{\mathit{Pol}}$ induces a Markov chain $\textbf{M}_\textbf{N}^\mathit{\overline{{Pol}}}$ when it is applied over the MDP $\textbf{M}_\textbf{N}$. This Markov chain is a disjoint union of a set of transient states $T_{\overline{\mathit{Pol}}}$ and $n$ sets of irreducible recurrent classes $R^k_{\overline{\mathit{Pol}}}$ \cite{stochastic} as: 

$$\textbf{M}_\textbf{N}^\mathit{\overline{{Pol}}}=T_{\overline{\mathit{Pol}}} \sqcup R^1_{\overline{\mathit{Pol}}} \sqcup ... \sqcup R^n_{\overline{\mathit{Pol}}}.$$ 
From (\ref{acc}), policy $\overline{\mathit{Pol}}$ satisfies $\varphi$ if and only if:
\begin{equation}
\label{recurrent_class}
\exists R^i_{\overline{\mathit{Pol}}} ~\mbox{s.t.}~ \forall j\in\{1,...,f\},~{F}^\otimes_j \cap R^i_{\overline{\mathit{Pol}}} \neq \emptyset.
\end{equation} 
The recurrent classes that satisfy \eqref{recurrent_class} are called accepting. From the irreducibility of the recurrent class $R^i_{\overline{\mathit{Pol}}}$ we know that all the states in $R^i_{\overline{\mathit{Pol}}}$ communicate with each other and thus, once a trace ends up there all the accepting sets are going to be visited infinitely often. Therefore, from the definition of $ \mathds{A} $ and accepting frontier function (Definition \ref{frontier}), the agent receives a positive reward $ r_p $ for ever once it reaches an accepting recurrent class $R^i_{\overline{\mathit{Pol}}}$. 

There are two other possibilities for the rest of the recurrent classes that are not accepting. A non-accepting recurrent class, such as $ R^k_{\overline{\mathit{Pol}}} $, either
\begin{enumerate}
\item has no intersection with any accepting set $ F_j^\otimes $, i.e. $$\forall j \in \{1,...,f\},~ F_j^\otimes \cap R^k_{\overline{\mathit{Pol}}} = \emptyset$$
\item or has intersection with some of the accepting sets but not all of them, i.e. 
$$ \exists J \subset 2^{\{1,...,f\}}\setminus\{1,...,f\}~\mbox{s.t.}~\forall j \in J,~ F_j^\otimes \cap R^k_{\overline{\mathit{Pol}}} \neq \emptyset $$
\end{enumerate}

In the first case, the agent does not visit any accepting set in the recurrent class and the only likelihood of visiting the accepting sets is in the transient states $ T_{\overline{\mathit{Pol}}} $ (which is not possible in LDBAs given the fact that $ \mathcal{Q}_D $ is invariant). Hence, the agent is able to visit some accepting sets but not all of them. This means that in the update rule of the frontier accepting set $ \mathds{A} $ in Definition \ref{frontier}, the case where $ q\in F_j \wedge \mathds{A}=F_j $ will never happen since there exist always at least one accepting set that has no intersection with $ R^k_{\overline{\mathit{Pol}}} $. Therefore, after a limited number of times, no positive reward can be obtained. This is also true in the second case since there is at least one accepting set that cannot be visited after the trace reaches $ R^k_{\overline{\mathit{Pol}}} $ which blocks the reinitialisation of $ \mathds{A} $ in Definition \ref{frontier}. 

Recall Definition \ref{expectedut}, where the expected utility for the initial state $\bar{s} \in \mathcal{S}^\otimes$ is defined as:
$$
{U}^{\overline{\mathit{Pol}}}(\bar{s})=\mathds{E}^{\overline{\mathit{Pol}}} [\sum\limits_{n=0}^{\infty} \gamma^n~ R(S_n,\overline{\mathit{Pol}}(S_n))|S_0=\bar{s}].
$$
In both cases, for any arbitrary $ r_p>0 $ (and $ r_n=0 $), there always exists a $ \gamma $ such that the expected utility of a trace hitting $ R^i_{\overline{\mathit{Pol}}} $ with unlimited number of positive rewards, is higher than the expected utility of any other trace. In the following, by contradiction, we show that any optimal policy
${\mathit{Pol}}^*$ which optimizes the expected utility will satisfy the
property.
 
\noindent
\textbf{Contradiction Assumption:} Suppose that the optimal policy ${\mathit{Pol}}^*$ does not satisfy the property $\varphi$. 

\noindent
By this assumption:
\begin{equation}\label{contradiction}
\forall R^i_{{\mathit{Pol}}^*},~\exists j \in \{1,...,f\},~ F_j^\otimes \cap R^i_{{\mathit{Pol}}^*} = \emptyset.
\end{equation} 
As we discussed in case 1 and case 2 above, the accepting policy $ \overline{\mathit{Pol}} $ has a higher expected utility than that of the optimal policy $ {\mathit{Pol}}^* $ due to limited number of positive rewards in policy $ {\mathit{Pol}}^* $. This is, however, in direct contrast with Definition \ref{optimal_pol} and hence the supposition is false and the given statement in the theorem is true. \hfill $\lrcorner$ 

\clearpage
\section*{Appendix 2}
\textbf{Corollary \ref{thm2}}. If no policy in MDP~$\textbf{M}$ can be generated to satisfy the property $ \varphi $, LCRL is still able to produce the best policy that is closest to satisfying the given LTL formula. \\
\\
Proof:\\
\noindent
Assume that there exists no policy in MDP~$\textbf{M}$ that can satisfy the property $\varphi$. Construct the induced Markov chain $\textbf{M}_\textbf{N}^\mathit{{{Pol}}}$ for any arbitrary policy ${\mathit{Pol}}$ and its associated  set of transient states $T_{{\mathit{Pol}}}$ and $n$ sets of irreducible recurrent classes $R^k_{{\mathit{Pol}}}$:
$$\textbf{M}_\textbf{N}^\mathit{{{Pol}}}=T_{{\mathit{Pol}}} \sqcup R^1_{{\mathit{Pol}}} \sqcup ... \sqcup R^n_{{\mathit{Pol}}}.$$ 
By the assumption of corollary, policy ${\mathit{Pol}}$ cannot satisfy the property and we have:
\begin{equation}
\forall R^i_{{\mathit{Pol}}},~\exists j \in \{1,...,f\},~ F_j^\otimes \cap R^i_{{\mathit{Pol}}} = \emptyset,
\end{equation} 
which essentially means that there are some automaton accepting sets like $ F_j $ that cannot be visited. Therefore, after a limited number of times no positive reward is given by the reward function $ R(s^\otimes,a) $. However, the closest recurrent class to satisfying the property is the one that intersects with more accepting sets.

By Definition \ref{expectedut}, for any arbitrary $ r_p>0 $ (and $ r_n=0 $), the expected utility at the initial state for a trace with highest number of intersections with accepting sets is maximum among other traces. Hence, by the convergence guarantees of QL, the optimal policy produced by LCRL converges to a policy whose recurrent classes have the highest number of intersections with automaton accepting sets. \hfill $ \lrcorner $

\clearpage
\section*{Appendix 3}
\textbf{Theorem \ref{thm1}}. If the LTL property is satisfiable by MDP~$\textbf{M}$, then the optimal policy generated by LCRL, maximizes the probability of property satisfaction.\\
\\
Proof:\\
\noindent
Assume that MDP $\textbf{M}_\textbf{N}$ is the product of an MDP~$\textbf{M}$ and an
automaton $\textbf{N}$ where $\textbf{N}$ is the automaton associated with the
desired LTL property $\varphi$. In MDP~$\textbf{M}_\textbf{N}$, a directed graph induced by a pair $({S^\otimes},A),~ S^\otimes\subseteq\mathcal{S}^\otimes,~A\subseteq\mathcal{A}$ is a Maximal End Component (MEC) if it is strongly connected and there exists no $({S^\otimes}',A')$ such that $({S^\otimes},A)\neq ({S^\otimes}',A')$ and ${S^\otimes} \subset {S^\otimes}'$ and $A \subset A'$ for all $s^\otimes \in {S^\otimes}$ \cite{bible}. A MEC is accepting if it contains accepting conditions of the automaton associated with the property $\varphi$. The set of all accepting MECs are denoted by \textit{AMECs}.

 Traditionally, when the MDP graph and transition probabilities are known, the probability of property satisfaction is often calculated via DP-based methods such as standard value iteration over the product MDP $\textbf{M}_\textbf{N}$ \cite{bible}. This allows us to convert the satisfaction problem into a reachability problem. The goal in this reachability problem is to find the maximum (or minimum) probability of reaching \textit{AMECs}. 

The value function $ V:\mathcal{S}^\otimes\rightarrow [0,1] $ is then initialised to $ 0 $ for non-accepting MECs and to $ 1 $ for the rest of the MDP. Once value iteration converges then at any given state $ s^\otimes=(s,q)\in\mathcal{S}^\otimes $ the optimal policy $ \pi^*:\mathcal{S}^\otimes\rightarrow\mathcal{A} $ is produced by
\begin{equation}\label{vi}
\pi^*({s^\otimes})=\arg\max\limits_{a}\sum\limits_{{s^\otimes}'\in\mathcal{S}} P({s^\otimes},a,{s^\otimes}')V^*({s^\otimes}'),
\end{equation}
where $ V^*$ is the converged value function, representing the maximum probability of satisfying the property at state $ s $. In the following we show that the optimal policy $ \mathit{Pol}^* $, generated by LCRL, is indeed equivalent to $ \pi^* $. 

They key to compare standard model-checking methods to LCRL is reduction of value iteration to basic form. Interestingly, quantitative model-checking over an MDP with a reachability predicate can be converted to a model-checking problem with an equivalent reward predicate which is called the basic form. This reduction is done by adding a one-off reward of $ 1 $ upon reaching \textit{AMECs} \cite{pareto}. Once this reduction is done, Bellman operation is applied over the value function (which represents the satisfaction probability) and policy $ \pi^* $ maximises the probability of satisfying the property. 

In LCRL, when an AMEC is reached, all of the automaton accepting sets will surely be visited by policy $ \mathit{Pol}^* $ and an infinite number of positive rewards $ r_p $ will be given to the agent (Theorem \ref{thm}). 

There are two natural ways to define the total discounted rewards \cite{dis2undis}:
\begin{enumerate}
\item to interpret discounting as the coefficient in front of the reward.
\item to define the total discounted rewards as a terminal reward after which no reward is given.
\end{enumerate}
It is well-known that the expected total discounted rewards corresponding to these methods are the same \cite{dis2undis}. Therefore, without loss of generality, given any discount factor $ \gamma $, and any positive reward component $ r_p $, the expected discounted utility for the discounted case (LCRL) is $ c>0 $ times the undiscounted case (value iteration). This means maximising one is equivalent to maximising the other.\hfill $ \lrcorner $

\end{document}